\documentclass[letterpaper]{article} 
\usepackage{aaai2026}  
\usepackage{times}  
\usepackage{helvet}  
\usepackage{courier}  
\usepackage[hyphens]{url}  
\usepackage{graphicx} 
\usepackage{natbib}  
\usepackage{caption} 
\usepackage{algorithmic}

\usepackage[ruled]{algorithm2e}

\usepackage{amsmath}
\usepackage{amsthm,amsmath,amssymb}
\usepackage{mathrsfs}

\usepackage{newfloat}
\usepackage{listings}
\DeclareCaptionStyle{ruled}{labelfont=normalfont,labelsep=colon,strut=off} 
\usepackage{multirow}
\usepackage{siunitx}
\usepackage[table]{xcolor}

\usepackage{colortbl}
\usepackage{float}
\usepackage{microtype}
\usepackage{graphicx}
\usepackage{subfigure}
\usepackage{multirow}
\usepackage{booktabs}
\usepackage{arydshln}

\usepackage{tikz}

\usepackage{makecell}

\usepackage{times}
\usepackage{helvet}
\usepackage{courier}
\usepackage{xcolor}

\usetikzlibrary{tikzmark,decorations.pathreplacing,calc}

\title{FLRQ: Faster LLM Quantization with Flexible Low-Rank Matrix Sketching}

\author {
    Hongyaoxing Gu\textsuperscript{\rm 1,\rm 2},
    Lijuan Hu\textsuperscript{\rm 1},
    Shuzi Niu\textsuperscript{\rm 1},
    Fangfang Liu\textsuperscript{\rm 1,3}\thanks{Corresponding author}
}
\affiliations {
    \textsuperscript{\rm 1}Institute of Software Chinese Academy of Sciences\\
    \textsuperscript{\rm 2}University of Chinese Academy of Sciences
\\
    \textsuperscript{\rm 3}Key Laboratory of System Software (Chinese Academy of Sciences)
}

\usepackage{bibentry}

\begin{document}

\maketitle

\begin{abstract}
Traditional post-training quantization (PTQ) is considered an effective approach to reduce model size and accelerate inference of large-scale language models (LLMs). However, existing low-rank PTQ methods require costly fine-tuning to determine a compromise rank for diverse data and layers in large models, failing to exploit their full potential. Additionally, the current SVD-based low-rank approximation compounds the computational overhead. In this work, we thoroughly analyze the varying effectiveness of low-rank approximation across different layers in representative models. Accordingly, we introduce \underline{F}lexible \underline{L}ow-\underline{R}ank \underline{Q}uantization (FLRQ), a novel solution designed to quickly identify the accuracy-optimal ranks and aggregate them to achieve minimal storage combinations. FLRQ comprises two powerful components, Rank1-Sketch-based Flexible Rank Selection (R1-FLR) and Best Low-rank Approximation under Clipping (BLC). R1-FLR applies the R1-Sketch with Gaussian projection for the fast low-rank approximation, enabling outlier-aware rank extraction for each layer. Meanwhile, BLC aims at minimizing the low-rank quantization error under the scaling and clipping strategy through an iterative method. FLRQ demonstrates strong effectiveness and robustness in comprehensive experiments, achieving state-of-the-art performance in both quantization quality and algorithm efficiency.

\end{abstract}

\begin{links}
\end{links}

\section{Introduction}
Large language models (LLMs) have demonstrated outstanding capabilities across a wide range of tasks that significantly impact our daily lives. However, the scale of LLMs makes their deployment extremely challenging, necessitating quantization techniques for reducing model size while improving inference efficiency \cite{kuzmin2023pruning}.

In quantization methods, Quantization-Aware Training (QAT) incorporates pseudo-quantization layers to simulate the quantization process \cite{liu-etal-2024-llm}, requiring retraining with different quantization parameters. This approach inevitably demands higher computational costs and can be affected by modifications to the original model architecture. In contrast, Post-Training Quantization (PTQ) has recently gained more attention for a faster quantization process and state-of-the-art PTQ methods have achieved accuracy comparable to QAT \cite{ding2022towards, hubara2021accurate, frantar2023gptq}. Among post-training quantization (PTQ) methods, low-rank approaches have garnered significant attention in recent years due to their effectiveness in capturing essential weight information, thereby enhancing quantization accuracy.
\begin{figure*}[t]
\centering
\includegraphics[width=1\linewidth]{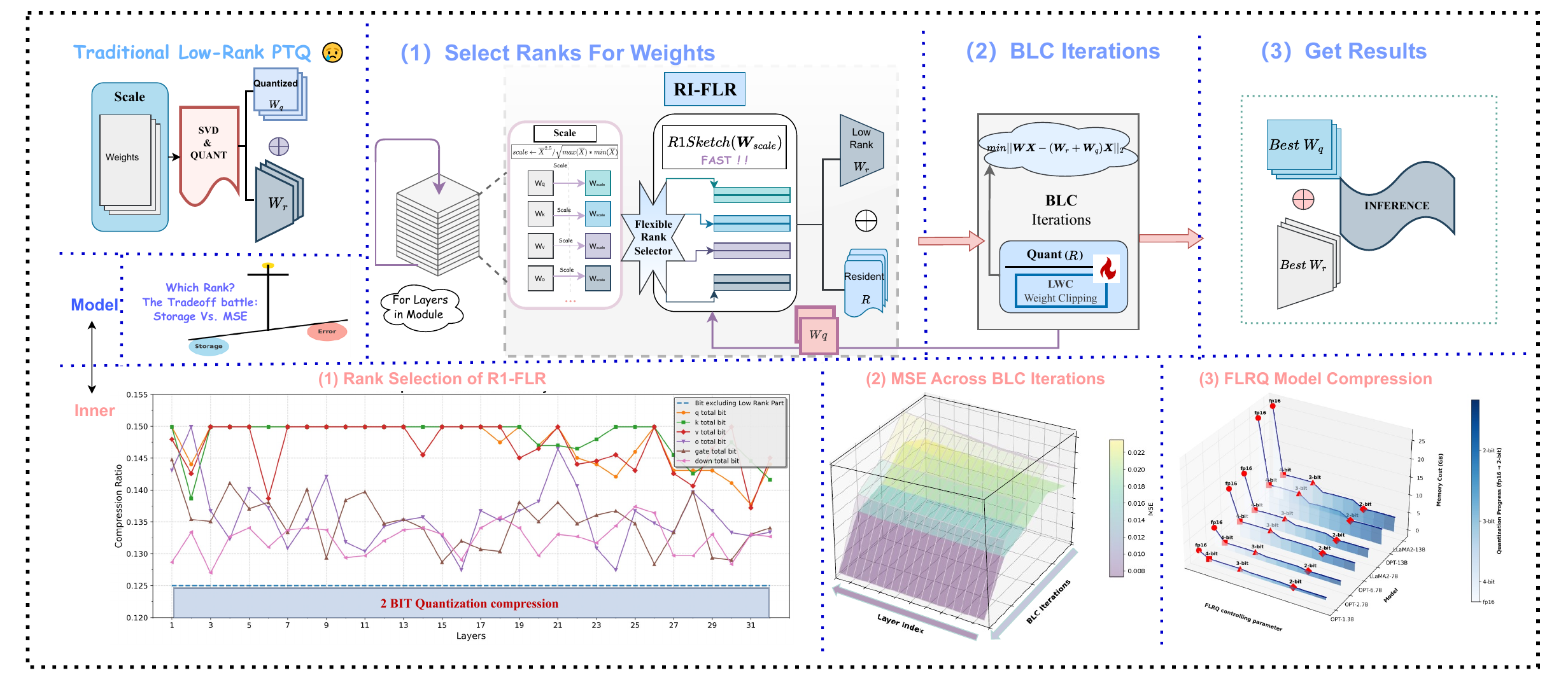}
\caption{The presentation at the model layer and internals of FLRQ, (1) to (3) represent the three steps of the FLRQ algorithm respectively. In detail, FLRQ utilizes flexible rank selections (1), activation-based scaling, and iterative Best Low-rank Approximation under Clipping (BLC) algorithm (2), leading to higher quantization accuracy and smaller model size (3). }
\label{fig_flrq}
\end{figure*}
\begin{table}[t]
    \small
  \centering
    \begin{tabular}{ccc}
    \toprule
    Method & Rank & Lora func \\
    \midrule
    LoRC\cite{yao2024exploring}    & 1,4,8,16,32 & SVD \\
    LoftQ\cite{liloftq} & 16,32 & SVD \\
    LQER\cite{zhang2024lqer}   & 600 & SVD \\
    L$^2$QER\cite{zhang2024lqer}  & 32,64,256 & SVD \\
    SVD-Quant\cite{li2024svdqunat}  & 16,32,64 & SVD \\
    \hdashline
    \rowcolor{blue!10} FLRQ(ours) & Flexible & R1-Sketch \\
    \bottomrule
    \end{tabular}%
  \caption{Previous low-rank quantization methods are constrained to restricted low-rank components, while FLRQ under R1-Sketch provides the flexibility to select effective low-rank approximations, thereby minimizing computational resource wastage and achieving high efficiency.}
  \label{method_rank}%
\end{table}

Current low-rank quantization approaches typically adopt a fixed rank and apply it uniformly across all layers of a model. However, recent studies reveal that different layers exhibit varying degrees of importance. This uniform-rank strategy has two primary drawbacks: 1. For models of different sizes, the memory usage ratio under the same rank varies, necessitating separate tuning of the fixed rank for each model; 2. A fixed rank leads to inconsistent effectiveness in capturing weight feature across layers, resulting in potential memory waste and insufficient performance.
 
Given these issues, we have proposed an iterative FLRQ method for flexible rank selection in different layers of the model, which is also shown in Figure \ref{fig_flrq}. Furthermore, we enhance the matrix sketching algorithm to accelerate our proposed method. Our key contributions are as follows.
\begin{itemize}
    \item We analyzed the effectiveness of reducing error in different layers of large model weights under low-rank approximation, and proposed FLRQ for selecting optimal low-rank components across various layers, which does not require expensive knowledge distillation, hyperparameter search, and is easily integrated with other approaches.
    \item We proposed rank-1 sketch (R1-Sketch) under Gaussian projection by reformulating the randomized SVD algorithm and carrying efficient implementation on GPU platform. This technique is suited for low-rank component extraction and solely utilizes BLAS Level-2 routines, ensuring highly efficient computations.
    \item Experiments demonstrate that FLRQ achieves comparable accuracy with fewer memory costs and outperforms conventional schemes by a significant margin in accuracy, especially for low-bit precision (INT2). Moreover, the R1-Sketch method effectively reduces computation time and has low inference latency.

\end{itemize}

\section{Background and Related Work}
\subsection{Post-Training Quantization}
PTQ (Post-Training Quantization), as a technique for model compression, has received extensive attention in recent years: Weight-only quantization in PTQ can be expressed as
\begin{equation}
\boldsymbol O = Quant(\boldsymbol W) \cdot \boldsymbol X.
\label{Weight-only-qgemm}
\end{equation}
AWQ \cite{lin2024awq} enhances the quantization accuracy by preserving the top 1\% of significant activation channels by applying a scale matrix. GPTQ \cite{frantar2023gptq} quantizes each parameter by OBS \cite{OBS} per block to mitigate the accuracy loss from quantization. 

Recent studies like Omniquant \cite{shao2023omniquant} introduced learnable weight clipping, Quip\# \cite{tseng2024quip} employes a randomized matrix transformation, while Affinequant \cite{maaffinequant} uses equivalent affine transformations, CALDERA \cite{saha2024compressing} applies low-rank decomposition in quantization. These methods consider the challenging regime of sub-4 bit post-training LLM quantization but there are still challenges for 2-bit quantization.

\subsection{Low-Rank methods in LLMs quantization}
Low-rank methods in quantization have attracted considerable attention in recent years. Low-rank quantization takes advantage of the low-rank decomposition of the weight quantization matrix \cite{yao2024exploring}, followed by residual quantization represented as:
\begin{equation}
    \boldsymbol W\boldsymbol X = (Quant(\boldsymbol W-\boldsymbol W_r) + \boldsymbol W_{r}) \boldsymbol  X,
\label{lorawx}
\end{equation}
where thr rank $r$ matrix \(\boldsymbol W_r \) is the low-rank approximation of \( W \) with minimal error, which can be calculated by SVD (Singular value decomposition):
\begin{equation}
    \boldsymbol W_r = SVD(\boldsymbol W)=(U_r \Sigma_r) V_r^T =\boldsymbol W_L \boldsymbol W_R.
\label{r-svd}
\end{equation}
A small subset of high singular value components can be isolated, and extracting these low-rank matrices effectively reduces the impact of outliers. This approach is broadly categorized into two types: 
\begin{itemize}
    \item \textbf{low-rank within quantization:} The former belongs to post-training quantization (PTQ) methods, where low-rank approximation is applied during quantization to reduce weight outliers and thereby minimize quantization error. For instance, ViTALiTy \cite{dass2023vitality} employs a combination of sparsification and low-rank approximation, while LQER \cite{zhang2024lqer} integrates Block Floating Point techniques to further enhance accuracy. Similarly, SVD-Quant \cite{li2024svdqunat} introduces low-rank methods into diffusion models, yielding significant performance improvements.

    \item \textbf{low-rank fine-tuning after quantization:} This involves refining the PTQ quantized model on a dataset to improve accuracy, as exemplified by methods such as LoftQ \cite{liloftq} achieves nearly lossless inference quantization by designing tailored singular value distributions. Recent work like CALDERA \cite{saha2024compressing} improves L2-loss by iteration and applies mix-precision in $\boldsymbol W_r$, and RILQ \cite{lee2025rilq} considers the construct of $\boldsymbol W_r$ by model loss instead of linear loss, both methods perform low-rank fine-tuning based on state-of-the-art PTQ like Omniquant \cite{shao2023omniquant} and achieve high-precision 2-bit models on QUIP\# \cite{tseng2024quip}.
\end{itemize}

\subsection{Sketching matrix and RSVD algorithm}
\label{section_sketch}
Matrix sketching is a technique designed to efficiently handle large-scale matrices. Its core idea is to construct a ``sketch'' through randomization. This approach reduces storage and computational costs while preserving most of the essential information. Randomized Singular Value Decomposition (RSVD) \cite{halko2011finding} represents one class of matrix sketching techniques utilized for low-rank approximation and has been widely applied in computer vision ~\cite{ji2014gpu, osawa2017accelerating} and machine learning~\cite{guan2017matrix, kumar2016novel}.
This algorithm is generally divided into the following two steps:
\begin{enumerate}
    \item Compute an approximate basis for the column space of \( A \in \mathbb{R}^{m \times n} \), assume $m\leq n$. Attempt it in $it$ times to obtain a matrix \( Q \) with \( r \) orthogonal columns that approximates matrix \( A \). Formally, \( A_r \approx QQ^*A \), where \( Q^* \) denotes the conjugate transpose of \( Q \).
    \item Utilize the orthogonal matrix \( Q \) to calculate a much smaller rank-\( k \) matrix \( Q^*A \), and employ it to compute the desired low-rank matrix by Singular Value Decomposition.
\end{enumerate}

It is evident that the computational requirements for RSVD are significantly reduced compared to SVD and the error satisfies \footnote{The proof of this theorem is complex; for specifics, one may refer to \cite{halko2011finding}.}:
\begin{equation}
\label{rsvderror}
\mathbb E \Vert A-A_r\Vert \leq \sigma_{r+1}+\left[ 1+4\sqrt{\frac{2n}{r-1} }\right ]^{1/(it+1)}\sigma_{r+1}.
\end{equation}
where $\sigma_{i}$ represents the $i$-th largest singular value of $A$.

\subsection{Motivation: Why do we need flexible and fast rank selection?}
In Low-rank methods, rank selection often relies on empirical insights in specific layers, and the selected rank is a fixed value for all weights. For example, LoRC in ZeroQuantv2 \cite{yao2024exploring} fixes the rank as \( 2^n \), RILQ and LoftQ use fixed ranks of 16 or 32, and experiments in LQER suggest that a rank of 64 is sufficient to maintain the accuracy of the OPT-1.3b model, SVD-Quant adopts a fixed rank of 32 while CALDERA applies ranks from 64 to 256. 


A fixed rank introduces non-negligible memory and latency costs. Under 2-bit quantization, LQER with rank 256 incurs roughly \textbf{50\%} additional memory, whereas CALDERA with rank 256, despite achieving higher accuracy, it brings 20\% extra memory and increasing inference latency by nearly \textbf{30\%} on 7B model. The overhead of SVD or fine-tuning–based quantization is also substantial: RILQ demonstrates competitive accuracy at rank 16 yet must be combined with other quantization methods—e.g., RILQ+OmniQuant requires 3.1 h for quantization and about 1 h for fine-tuning on LLaMA2-7B and SVD-based methods severely slow quantization, diminishing usability.
\section{Method}
\label{FLRQ}
In quantization, low-rank methods frequently employ a fixed rank approach, which is not ideally suited for all layers. We introduce \textbf{FLRQ} (\underline{F}lexible \underline{L}ow-\underline{R}ank \underline{Q}uantization), structured into two primary components and presented in Algorithm \ref{algorithmofFLRQ}:
\begin{itemize}
    \item \textbf{R1-FLR} (R1-Sketch-based Flexible Rank Selection): This component focuses on dynamically determining the rank for each layer under rank-1 sketch (R1-Sketch), allowing for adaptability according to the specific characteristics and requirements of individual layers.
    \item \textbf{BLC} (Best Low-rank Approximation under Clipping): This segment aims at achieving efficient low-rank approximations when utilizing clipping strategies to minimize quantization errors while preserving critical data features.
\end{itemize}
In this section, we will first present the R1-Sketch algorithm, followed by the introduction of two optimization methods: \textbf{R1-FLR} and \textbf{BLC}.

\subsection{Low rank under R1-Sketch}
\label{r1-sketch-section}
To address the inefficiency of SVD, we proposed a simplification to the randomized SVD (RSVD) algorithm under the rank-1 condition. This simplification introduces a rank-1 matrix approximation technique, as described in below:

Given a matrix $\boldsymbol A \in \mathbb{R}^{m\times n}$. For a standard RSVD (Randomized Singular Value Decomposition) algorithm prototype, it typically consists of the following two steps:\\
\textbf{Stage A:}
\begin{enumerate}
    \item Generate an $\mathbb{R}^{n\times r}$ Gaussian test matrix $\boldsymbol S$.
    \item Form $\boldsymbol Y=(\boldsymbol A\boldsymbol A^{*})^{it}\boldsymbol A \boldsymbol S$.
    \item Construct a matrix $\boldsymbol Q = QR(\boldsymbol Y)$ by QR decomposition whose columns form an orthonormal basis of $\boldsymbol Y$.
\end{enumerate}
\textbf{Stage B:}
\begin{enumerate}
    \item Form $\boldsymbol B = \boldsymbol Q^{*}\boldsymbol A$.
    \item Compute an SVD of the small matrix: $\boldsymbol B=\boldsymbol U\boldsymbol \Sigma \boldsymbol V^{*}$.
    \item Set $\boldsymbol U= \boldsymbol Q \boldsymbol U$.
\end{enumerate}

If a rank-1 matrix \( \boldsymbol S\in \mathbb{R}^{n\times1} \) is utilized for low-rank approximation of a matrix, substitute the rank-1 matrix in the two stages, we also have $Y = (\boldsymbol A \boldsymbol A^{*})^{it} \boldsymbol A  S$.

then for the matrix \(\boldsymbol Y \in \mathbb{R}^{m \times 1} \) , the QR decomposition can be directly represented as follows. 
\begin{equation}
    \boldsymbol Q = \frac{\boldsymbol  Y}{\Vert  \boldsymbol  Y \Vert} \in \mathbb{R}^{m \times 1} , \boldsymbol  R = \Vert  \boldsymbol Y \Vert \in \mathbb{R}^{1 \times 1}.
\label{Rank1QR}
\end{equation}
Similarly, the SVD decomposition for rank-1 matrix \( \boldsymbol B = \boldsymbol Q^* \boldsymbol A \) can be represented as follows:
\begin{equation}
   \boldsymbol  U = \{1\},\boldsymbol  \Sigma = \Vert \boldsymbol B \Vert, \boldsymbol V = \frac{\boldsymbol B}{\Vert\boldsymbol  B \Vert}.
\label{Rank1SVD}
\end{equation}
Denote \( A_L =\boldsymbol Q \boldsymbol U\boldsymbol \Sigma \) and \( A_R = \boldsymbol V \). Apply the results of SVD decomposition, we have:
\begin{equation}
\label{Lowrank_ALAR}
\begin{aligned}
  A_L &=  \frac{(\boldsymbol A\boldsymbol A^*)^{it}\boldsymbol A\boldsymbol S}{\Vert(\boldsymbol A\boldsymbol A^*)^{it}\boldsymbol A\boldsymbol S\Vert} \cdot \frac{\Vert \boldsymbol S^*\boldsymbol A^*(\boldsymbol A\boldsymbol A^*)^{it}\boldsymbol A \Vert}{\Vert(\boldsymbol A\boldsymbol A^*)^{it}\boldsymbol A\boldsymbol S\Vert}, \\
  A_R &= \frac{\boldsymbol B}{\Vert \boldsymbol B\Vert} = \frac{ \boldsymbol S^*\boldsymbol A^*(\boldsymbol A\boldsymbol A^*)^{it}\boldsymbol A }{\Vert \boldsymbol S^*\boldsymbol A^*(\boldsymbol A\boldsymbol A^*)^{it}\boldsymbol A \Vert}.
\end{aligned}
\end{equation}
\subsection{R1-Sketch-based Flexible Low-Rank Selection}
Given a layer $\mathbf{W}\in \mathbb{R}^ {m*n}$ in $d_{fp}$ and perform a pseudo-quantization with $r$ rank $\mathbf{W_r}$ under $d$ bit.
\begin{equation}
\label{calERands}
\widehat{\boldsymbol W_q} = clamp(\lfloor \boldsymbol  R/s_r \rceil)*s_r  ;  \ s_r = \frac{2^{d-1} \text{-}  1}{amax(\boldsymbol R)},
\end{equation}

where $\boldsymbol R = \boldsymbol W \text{-} \boldsymbol W_r $. Then $\widehat {\boldsymbol W}$ can be decomposed into $ \widehat {\boldsymbol W} = \widehat {\boldsymbol W_q }+ \boldsymbol W_r$. The low-rank component is stored in original precision which is nearly lossless. 

\begin{figure}[t]
\centering
\includegraphics[width=1\linewidth]{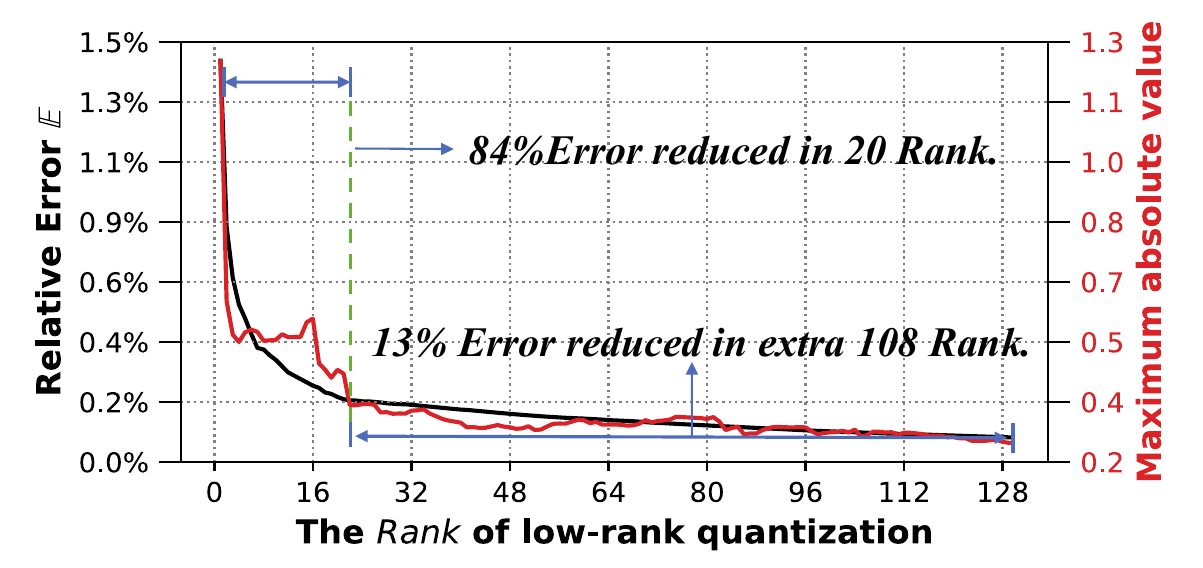}
\caption{ As the extraction rank increases, the curves of the relative error $\mathbb{E}$ under $L_2$ norm and the $amax$ value decrease. }
\label{fig_mse_abs}
\end{figure}

As a example illustrated in Figure \ref{fig_mse_abs}, the quantization error $\mathbb{E}=\Vert \boldsymbol W \boldsymbol  X - \widehat{\boldsymbol W}\boldsymbol  X \Vert /\Vert \boldsymbol W \boldsymbol X \Vert$ 
gradually decreases as the rank increases. However, repeated recalculation of $\mathbb{E}$ after layer re-quantization following rank adjustments incurs significant computational costs. In order to improve quantization efficiency, we propose a flexible rank selection approach based on $amax$ aimed at minimizing quantization errors while optimizing computational efficiency.

Assume that the corresponding \textit{amax} for this rank is \( w_r \). Then the maximum quantization error under rank $r$ quantization is $E_r=\frac{1}{2s_r}$. Thus, the error reduced by \( p =  \frac{E_0}{E_r} = \frac{w_0}{w_r}\), the precision improvement corresponding to an additional \( d' = \log_2(p) \) bits. Consequently, the model bit-width increases to \( d + d' \), and the effective quantization bits increase by $q$ in Eq.\ref{calQK}; For memory costs, after incorporating rank-$r$ matrix, the model size increased by $k$ in Eq.\ref{calQK}.
\begin{equation}
\label{calQK}
q = \frac{d + d'}{d} , k = 1 + \frac{d_{fp} \times r \times (m + n)}{d\times m\times n}.
\end{equation}
Then, \textbf{R1-FLR} workflow is as follows, and shown in Algorithm \ref{algorithmofR1FLR}:

Starting the traversal of $r$ from 1, applying the R1-Sketch method for computing the low-rank matrix corresponding to the current rank-1 matrix, then utilizing Equation \ref{calQK} to calculate the parameters q and k during each iteration. 

\begin{enumerate}
    \item If \( q > k \), it demonstrates that the precision improvement achieved exceeds the increase in model size. Thus, the r-rank quantization is effective.
    \item If \( q \leq k \), it indicates that the precision improvement achieved is not larger than the increase in model size due to the extraction of low-rank components.
\end{enumerate}

\begin{algorithm}[t]
   \caption{R1-Sketch Flexible Low-Rank Selection (R1-FLR)}
   \label{algorithmofR1FLR}
  \SetAlgoLined
\KwData{$W \in \mathbb{R}^{m*n}$(weight), $X$(value from calibration),)}
  
  \KwResult{$W_r, W_q$}

    Begin \textbf{R1-FLR} processes: \;
    \For{$i=1$ {\bfseries to} $n$}{
     Obtain the sketch rank-1 matrix by Eq.\ref{Lowrank_ALAR}: $\{\boldsymbol U_1,  \boldsymbol V_1\} \leftarrow calR1matrix(\boldsymbol R)$ \;
     $\boldsymbol R \leftarrow  \boldsymbol R - \boldsymbol U_1\cdot \boldsymbol V_1$\;
     get amax now: $maxAbs \leftarrow  amax(\boldsymbol R)$\;
     Calculate Q, K by Eq.\ref{calQK} and the slope:
     $sNow \leftarrow  getSlope(maxAbs)$\;
    \If{$K > Q \ or \ K > 1+x \ or\ sNow < t$}{
         Endloop\;
    }
     $\boldsymbol W_L$.append($U_1$); $\boldsymbol W_R$.append($V_1$)\;
    }
   \Return $\boldsymbol W_L, \boldsymbol W_R$\;
\end{algorithm}

\subsection{Best Low-rank Approximation under Clipping}
\textbf{Low-rank quantization with calibration}: Recent studies have shown that the data distribution of activation can influence the effectiveness of quantization \citet{lin2024awq, zhang2024lqer}. Scaling the weights according to the activation values can significantly reduce quantization errors. Specifically, this involves using a calibration dataset to perform inference on each layer during quantization to obtain the current layer's activation value distribution, then calculating a scale vector that is applied before low-rank approximation, which can perform as:
\begin{equation}
\label{scale}
    \{\boldsymbol U',\boldsymbol V\} = \text{R1-FLR}(\alpha\boldsymbol W), \ \ {\boldsymbol U} = \alpha^{-1}\boldsymbol U'.
\end{equation}
The calculation of $\alpha$ is:
\begin{equation}
  \alpha = {\overline {\boldsymbol X}}^{2.5}/\sqrt{max(\overline {\boldsymbol X})*min(\overline {\boldsymbol X})},
\label{calSfrommean}
\end{equation}
where ${\boldsymbol X}$ is the corresponding activation value of $\boldsymbol W$ and $\overline {\boldsymbol X}$ is per-token normalized mean of ${\boldsymbol X}$. The calculation here is similar to that in AWQ \cite{lin2024awq}.
In addition, during quantization, setting a portion of the numbers with the largest absolute values to zero by clipping can improve quantization accuracy.

\begin{table*}
  \centering

  \small
  \renewcommand{\arraystretch}{1.2}
    \begin{tabular}{c|c|c|c|c|c|c|c|c|c|c|c}
    \Xhline{4\arrayrulewidth}
    \multirow{2}[4]{*}{Precision} & \multirow{2}[4]{*}{Method} & \multicolumn{2}{c|}{OPT-1.3b} & \multicolumn{2}{c|}{OPT-6.7b} & \multicolumn{2}{c|}{OPT-13b} & \multicolumn{2}{c|}{LLaMA2-7b} & \multicolumn{2}{c}{LLaMA2-13b} \\
\cline{3-12}          &       & Wiki  & C4    & Wiki  & C4    & Wiki  & C4    & Wiki  & C4    & Wiki  & C4 \\
    \Xhline{3\arrayrulewidth}
    FP16  & Baseline & 14.62  & 14.72  & 10.86  & 11.74  & 10.13  & 11.19  & 5.47  & 6.97  & 4.88  & 6.47  \\
    \hline
    \multirow{5}[2]{*}{W4A16} & RTN   & 31.96  & 21.45  & 12.05  & 13.37  & 11.41  & 12.41  & 5.88  & 7.30  & 5.12  & 6.60  \\
          & AWQ   & 15.22  & 15.04  & 10.93  & 11.87  & 10.21  & 11.28  & 5.61  & 7.13  & 4.97  & 6.56  \\
          & OmniQuant & 14.88  & 15.03 & 10.96  & 11.85 & 10.20  & 11.29 & 5.58  & 7.12  & 4.95  & 6.56 \\
          & AffineQuant & 14.79 & 14.98 & 10.92 & 11.84 & 10.19 & 11.27 & 5.58  & 7.12  & 4.95  & 6.56 \\
          \rowcolor{black!10} & FLRQ  & \textbf{14.65}$\downarrow$& \textbf{14.97}$\downarrow$& \textbf{10.84}$\downarrow$& \textbf{11.84}$\downarrow$& \textbf{10.13}$\downarrow$& \textbf{11.28}$\downarrow$& \textbf{5.55}$\downarrow$& \textbf{7.06}$\downarrow$& \textbf{4.94}$\downarrow$& \textbf{6.52}$\downarrow$\\
    \hline
    \multirow{5}[2]{*}{W3A16} & RTN   & 119.1   & 126.47 & 23.54 & 32.56 & 46.03 & 44.12 & 6.66  & 8.4   & 5.51  & 7.18 \\
          & AWQ   & 16.32 & 16.27 & 11.41 & 12.30  & 10.68 & 11.61 & 6.24  & 7.84  & 5.32  & 6.94 \\
          & OmniQuant & 15.72 & 16.11 & 11.27 & 12.31 & 10.47 & 11.63 & 6.03  & 7.75  & 5.28  & 6.98 \\
          & AffineQuant & 15.61 & \textbf{16.02} & 11.18 & \textbf{12.21} & 10.51 & 11.63 & 6.08 & 7.83  & 5.28  & 6.99 \\
          \rowcolor{black!10} & FLRQ  & \textbf{15.53}$\downarrow$& 16.07  & \textbf{11.18}$\downarrow$& 12.37  & \textbf{10.52}$\downarrow$& \textbf{11.68}$\downarrow$&\textbf{5.88}$\downarrow$ & \textbf{7.45}$\downarrow$& \textbf{5.16}$\downarrow$& \textbf{6.77 } $\downarrow$\\
    \hline
    \multirow{5}[2]{*}{W2A16} & RTN   & 1.3e4 & 7.7e3 & 7.8e3 & 5.2e3 & 7.6e4 & 2.8e4 & 4.2e3 & 4.9e3 & 122.2 & 139.6 \\
          & AWQ   & 47.97 & 38.40  & 16.2  & 16.48  & 14.32 & 14.73  & 2.2e5 & 1.7e5 & 1.2e5 & 9.4e4 \\
          & OmniQuant & 23.95 & 27.33 & 14.43 & 16.67 & 12.94 & 14.92 & 11.06 & 15.02 & 8.26  & 11.05 \\
          & AffineQuant & 23.32 & 23.28 & 14.18 & 15.62 & 12.88 & 14.60  & 10.87 & 13.13 & 7.64  & 10.32 \\
          \rowcolor{black!10} & FLRQ  & \textbf{22.99}$\downarrow$& \textbf{22.52}$\downarrow$& \textbf{14.05}$\downarrow$& \textbf{15.23}$\downarrow$& \textbf{12.60}$\downarrow$& \textbf{13.81}$\downarrow$& \textbf{9.14}$\downarrow$& \textbf{12.10}$\downarrow$& \textbf{6.77}$\downarrow$& \textbf{8.87}$\downarrow$\\
\Xhline{4\arrayrulewidth}
    \end{tabular}%
  \caption{WikiText2 and C4 perplexity (PPL $\downarrow$)  results on OPT and LLaMA-2 models, context length is 2048.}
  \label{PPL_WIKI_MAIN}%
\end{table*}%

\begin{algorithm}[t]
   \caption{Flexible Low-Rank Matrix Sketching Quantization}
   \label{algorithmofFLRQ}
  \SetAlgoLined
\KwData{$W \in \mathbb{R}^{m*n}$(weight), $X$(value from calibration)}
  
  \KwResult{$W_r, W_q$}
    Init quantization: $\boldsymbol W_r = SVD(\boldsymbol W)$, $ \boldsymbol W_q = Quant(\boldsymbol W - \boldsymbol W_r)$, $bestE \leftarrow inf$\;
    Start \textbf{BLC} processes\;
    \For{$i=1$ {\bfseries to} $epochs$}{
        $\mathbb{E} \leftarrow ||\boldsymbol W\boldsymbol X - (\boldsymbol W_r + \boldsymbol W_q)\boldsymbol X||_2$\;

        \If{$\mathbb{E}<minE$}{
            $minE \leftarrow \mathbb{E}$\;
            $\{bestW_r,bestW_q\} \leftarrow \{\boldsymbol W_r,\boldsymbol W_q\} $\;
        }        
        Calculate the resident matrix: $\boldsymbol R \leftarrow \boldsymbol W -\boldsymbol W_q$\;
        $\boldsymbol W_r =\{\boldsymbol W_L,\boldsymbol W_R \}\leftarrow R1\text{-}{FLR}(\boldsymbol R )$\;

        Apply weight clipping to rest of $\boldsymbol W$: $\boldsymbol W_{clp} \leftarrow Clipping(\boldsymbol W-\boldsymbol W_r, p_{clp})$\;
        Quantize the $W_{clip}$: $\boldsymbol W_q = Quant(\boldsymbol W_{clp})$\;

    }
   \Return $best\boldsymbol W_r, best\boldsymbol W_q$\;
\end{algorithm}

\textbf{Get best low-rank quantization by iteration}: With the aforementioned strategies, the algorithm flow is as follows: 
\begin{enumerate}
    \item Under calibration, use R1-FLR to compute the low-rank matrix with scaling to obtain $\boldsymbol W_r$.
    \item Apply clipping to find a $p_{clp}$ and cut off the elements whose absolute values exceed $p_{clp}$, where $\boldsymbol W_{clp} = Clipping(\boldsymbol W - \boldsymbol W_r, p_{clp})$.
    \item Quantize the clipped matrix $\boldsymbol W_q =Quant(\boldsymbol W_{clp})$.
\end{enumerate}
In this context, is such a decomposition optimal? Considering this from error under $L_2$ norm, we aim to find a $d$ bit quantized matrix $\boldsymbol W_q$ and a rank $r$ matrix $\boldsymbol W_r$ that minimize
\begin{equation}
\label{minerr}
\begin{split}
    \mathop{min}\limits_{r, p_{clp}} \bigg [\mathbb{E}||\boldsymbol W\boldsymbol X - (\boldsymbol W_r + \boldsymbol W_q)\boldsymbol X||_2 \bigg ].
\end{split}
\end{equation}
Since the original problem is non-trivial and difficult to solve directly, we propose an iterative method \textbf{BLC} to progressively compress the residual space between the low-rank quantization and the original weights, thereby obtaining a minimized error. The core strategy of \textbf{BLC} involves the alternating update of $\boldsymbol W_r$ and $p_{clp}$.
Firstly, we apply \textbf{R1-FLR} and clipping to obtain $\boldsymbol W_r$ and $\boldsymbol W_q$. Then loop through the following three operations below to find the $\boldsymbol W_q$ and $\boldsymbol W_r$ corresponding to the minimum error $\mathbb{E}$:
\begin{enumerate}
    \item Calculate $\mathbb{E} = ||\boldsymbol W\boldsymbol X - (\boldsymbol W_r + \boldsymbol W_q)\boldsymbol X||_2$.
    \item Calculate the quantized residual matrix $\mathbf{R} =\boldsymbol  W - \boldsymbol W_q$ and obtain $\boldsymbol W_r = \textbf{R1-FLR}(\mathbf{R})$.
    \item Apply \textbf{clipping} to find $p_{clp}'$ for clipping and $\boldsymbol W_q = Quant(Clipping(\boldsymbol W-\boldsymbol W_r,p_{clp}'))$.
    \item Update the $\boldsymbol W_q$, $\boldsymbol W_r$ corresponding to the minimum $\mathbb{E}$.
    
\end{enumerate}

Now we present FLRQ algorithm, which incorporates the BLC iteration and R1-FLR flexible rank selection and the pseudo-code is shown in Algorithm \ref{algorithmofFLRQ}.

\section{Experiments}
\textbf{SetUp:}
We performed a series of evaluations on FLRQ. During quantization, the hyperparameter $it$ at $\mathbf{2}$ in main evaluations and the group size was fixed at \textbf{128}, aligning with the settings in AWQ quantization. We employ a calibration dataset consisting of 128 randomly selected 2048 token segments from WikiText2 \cite{merity2016pointer}, 
 which proved to be a good sampling strategy in OminiQuant \cite{shao2023omniquant}. All experiments were conducted on an Nvidia A100 40G GPU. Our evaluations included perplexity and zero-shot tests in the OPT \cite{zhang2022OPT}, LLaMA-2 and  LLaMA-3 \cite{touvron2023llama}  model families for language generation tasks. 


\textbf{Baselines:}
We used AWQ \cite{lin2024awq}, LQER \cite{zhang2024lqer}, OmniQuant \cite{shao2023omniquant} and Affinequant \cite{maaffinequant} as the primary baselines for comparison.

\textbf{Evaluation Tasks:}
For the evaluation tasks, we conducted perplexity experiments on the WikiText2 \cite{merity2016pointer} and C4 \cite{raffel2020exploring} datasets. We performed zero-shot experiments using test sets including ARC(challenge,easy) \cite{boratko2018systematic}, BOOLQ \cite{clark2019BOOLQ}, OpenBook-QA \cite{Mihaylov_Clark_Khot_Sabharwal_2018}, PIQA \cite{Bisk_Zellers_Le_bras_Gao_Choi_2020}, and Winogrande \cite{sakaguchi2021winogrande}, with the lm-evaluation-harness \cite{evalharness} framework testing.

\subsection{Language Generation}
We tested FLRQ on model perplexity at 4,3,2-bit. The results are presented in Table \ref{PPL_WIKI_MAIN} and the memory costs are shown in Table \ref{tab_rank}. According to the results, FLRQ outperforms four PTQ methods on most tasks. Then we compared FLRQ with LQER and the results are presented in Table \ref{tab_lqer}. In LQER's 2-bit quantization, a significantly high fixed rank of 256 is required to maintain accuracy, whereas in FLRQ, the average rank at 2-bit is only around 40, achieving better perplexity. This is attributed to our flexible rank selection and BLC method, which reduces the residual space thereby making our approach superior to non-iterative methods with fixed ranks  as shown in Figure \ref{fig_flrq}.(2).

Recently, low-rank quantized fine-tuning algorithms have received widespread attention. We compare our approach with two fine-tuning methods, CALDERA and RILQ, both of which achieve their optimal implementations based on Quip\#. However, CALDERA, despite achieving the highest accuracy, incurs significant latency during inference due to its selection of a larger rank (256-rank in int4). Furthermore, low-rank methods generally underperform compared to rotation-based approaches such as Quip\#; nevertheless, with the incorporation of fine-tuning, they can achieve comparable accuracy, demonstrating the robustness of the FLRQ algorithm.
\begin{table}[t]
  \centering
  \setlength{\tabcolsep}{4pt}
  \small
    \renewcommand{\arraystretch}{1.2}

    \begin{tabular}{cccccc}
    \Xhline{4\arrayrulewidth}
    \multirow{2}[4]{*}{Bit} & \multicolumn{3}{c}{OPT} & \multicolumn{2}{c}{LLaMA2} \\
\cmidrule{2-6}          & 1.3B  & 6.7B  & 13B   & 7B    & 13B \\
    \Xhline{3\arrayrulewidth}
    4     & 30.5/0.34 & 27.1/0.16 & 27.0/0.12 & 36.1/0.21 & 38.6/0.18 \\
    3     & 28.8/0.33 & 27.7/0.16 & 26.4/0.12 & 35.8/0.21 & 38.4/0.18 \\
    2     & 27.6/0.33 & 32.7/0.19 & 33.6/0.15 & 39.2/0.24 & 41.9/0.20 \\
    \Xhline{4\arrayrulewidth}
    \end{tabular}%
  \caption{The extracted rank and extra average bit width of FLRQ at different $x$ values as (rank/extra-avg.bit).}
  \label{tab_rank}%
\end{table}%

\begin{table}[t]
  \centering
  \small
    \renewcommand{\arraystretch}{1.2}

    \begin{tabular}{c c c c c c}
    \Xhline{4\arrayrulewidth}
    Bit   & Method & extra.bit$\downarrow$& avg.rank$\downarrow$& Wiki2$\downarrow$& C4$\downarrow$\\
    \Xhline{3\arrayrulewidth}
    \multirow{2}[1]{*}{3} & LQER  & 0.21  & 32    & 6.23  & 8.82  \\
         & \cellcolor{black!10}FLRQ  & \cellcolor{black!10}0.23  & \cellcolor{black!10}36    & \cellcolor{black!10}\textbf{5.88}  & \cellcolor{black!10}\textbf{7.45}  \\
    \midrule
    \multirow{2}[1]{*}{2} & LQER  & 1.60  & 256   & 10.33  & 12.12  \\
          & \cellcolor{black!10}FLRQ  & \cellcolor{black!10}0.24  & \cellcolor{black!10}39    & \cellcolor{black!10}\textbf{9.14} & \cellcolor{black!10}\textbf{12.10}  \\
    \Xhline{4\arrayrulewidth}
    \end{tabular}%
  \caption{Compare with LQER on LLaMA2-7b.}

  \label{tab_lqer}%
\end{table}
\begin{table}[t]
  \centering
  \small
    \renewcommand{\arraystretch}{1.2}

    \begin{tabular}{c|c|c|cc|c}
    \Xhline{4\arrayrulewidth}
    
    \multirow{2}[2]{*}{Method} & avg.  & extra. & \multicolumn{2}{c|}{GEN Tasks $\downarrow$} & low-rank  \\
          & rank  & bit   & Wiki2 & C4    &  latency$\downarrow$\\
    \Xhline{3\arrayrulewidth}
    Quip\# & -     & -     & 12.74 & 16.84 & - \\
    \rowcolor{black!10} FLRQ  & 40    & 0.24  & 14.12  & 17.81 & 4.8\% \\
    \hdashline
    Quip\&CALD & 256   & 0.4   & 8.87  & 12.02 & 26.7\% \\
    Quip\&RILQ & 64    & 0.4   & 9.64  & 12.96 & 7.1\% \\
    \rowcolor{black!10} FLRQ\&RILQ & 56    & 0.36  &   9.78    &    13.03   & 6.5\% \\
    \Xhline{4\arrayrulewidth}
    \end{tabular}%
  \caption{2-bit PPL and inference latency on LLaMa3-8B, where CALD is short of CALDERA. QUIP\# and FLRQ are PTQ quantization methods, while RILQ and CALDERA are low-rank fine-tuning methods.}
  \label{tab:addlabel}%
\end{table}%

\subsection{Downstream task accuracy}
We reused the quantization configuration in language generation and performed a zero-shot evaluation on six downstream tasks, including ARC (easy), ARC (challenge), PIQA, OpenBookQA, BOOLQ, and Winogrande, and the results are presented in Table \ref{ZERO_shot_MAIN}. FLRQ's average accuracy across the six downstream tasks is consistent with the FP16 baseline, showing near-lossless accuracy in 4bit and 3bit quantization, and maintains considerable accuracy at 2 bits.

\subsection{The hyperparameters in FLRQ}
\textbf{The choice of iterations $it$}: The accuracy of r1-sketch is determined by the number of iterations $it$. We evaluated the convergence and execution efficiency under varying values of $it$ and the effectiveness of parameter selection $it$ in the FLRQ method. The evaluation concludes with PPL and execution time in Table \ref{tab_iter}. The findings indicate that while a higher $it$ improves the accuracy of R1-Sketch,  setting $it = 2$ in FLRQ is enough. In this case, R1-Sketch only takes \textbf{6} $GEMV$ of $O(N^2)$ and some $O(N)$ routines, providing a balance between accuracy and computational efficiency.

{\setlength{\tabcolsep}{5pt}
\begin{table}[t]
  \centering

  \small
    \renewcommand{\arraystretch}{1.2}
    \begin{tabular}{cccccc}
    \Xhline{4\arrayrulewidth}
    \multirow{2}[4]{*}{Method} & \multicolumn{3}{c}{OPT} & \multicolumn{2}{c}{Llama2} \\
\cmidrule{2-6}          & 1.3b  & 6.7b  & 13b   & 7b    & 13b \\
    \Xhline{3\arrayrulewidth}
    FP16  & 48.8\% & 55.2\% & 55.6\% & 62.7\% & 63.4\% \\
    \hdashline
    AWQ(4) & 47.4\% & 54.7\% & 55.7\% & 62.2\% & 64.1\% \\
    Omni(4) & 47.8\% & 54.9\% & 55.6\% & 62.5\% & 65.0\% \\
    \rowcolor{black!10} FLRQ(4) & \textbf{48.4\%}$\uparrow$ & \textbf{55.0\%}$\uparrow$ & \textbf{55.1\%}$\uparrow$ & \textbf{62.7\%}$\uparrow$ & \textbf{65.4\%}$\uparrow$ \\
    \hline
    AWQ(3) & 46.6\% & 53.2\% & 53.5\% & 60.9\% & 62.7\% \\
    Omni(3) & 47.4\% & 53.7\% & 54.8\% & 61.1\% & 63.4\% \\
    \rowcolor{black!10} FLRQ(3) & \textbf{47.6\%}$\uparrow$ & \textbf{54.4\%}$\uparrow$ & \textbf{55.3\%}$\uparrow$ & \textbf{61.4\%}$\uparrow$ & \textbf{64.2\%}$\uparrow$ \\
    \hline
    Omni(2) & 41.9\% & 46.9\% & 47.3\% & 50.1\% & 53.9\% \\
    \rowcolor{black!10} FLRQ(2) & \textbf{45.5\%}$\uparrow$ & \textbf{51.5\%}$\uparrow$ & \textbf{52.3\%}$\uparrow$ & \textbf{56.4\%}$\uparrow$ & \textbf{60.4\%}$\uparrow$ \\
    \Xhline{4\arrayrulewidth}
    \end{tabular}%
  \caption{Zero-Shot results ($\uparrow$) in an average across six zero-shot tasks. Omni is in short of Omniquant and the quantization bit is given in parentheses.}
  \label{ZERO_shot_MAIN}%
\end{table}}

\begin{table}[t]
  \centering
  \small
    \renewcommand{\arraystretch}{1.2}

    \begin{tabular}{ccccc}
    \Xhline{4\arrayrulewidth}
    \multirow{2}[4]{*}{it} & \multicolumn{2}{c}{OPT-1.3B} & \multicolumn{2}{c}{OPT-6.7B} \\
\cmidrule{2-5}          & PPL   & Time  & PPL   & Time \\
    \Xhline{3\arrayrulewidth}
    0     & 16.84 & 6.1m/1.2s & 11.65 & 26.5m/3.1s \\
    1     & 15.56 & 6.1m/1.8s & 11.42 & 26.5m/5.2s \\
    \rowcolor{black!10}  \textbf{2} & \textbf{15.53} & \textbf{6.1m/2.1s} & \textbf{11.18} & \textbf{26.6m/9.3s} \\
    4     & 15.53 & 6.1m/3.5s & 11.18 & 26.7m/16.9s \\
    8     & 15.53 & 6.2m/6.8s & 11.18 & 27.0m/29.5s \\
    \hline
    SVD   & 15.53 & 8.7m/2.6m & 11.18 & 56.4m/20.3m \\
    \Xhline{4\arrayrulewidth}
    \end{tabular}%
  \caption{The PPL and time costs (\textbf{FLRQ} total time)/(R1-FLR partial time) of FLRQ with different $it$ parameters on 3-bit quantized OPT models under wikitext2 dataset.}
  \label{tab_iter}%
\end{table}%

\subsection{Time efficiency}
\label{time_efficiency}
We evaluated the efficiency of FLRQ from two perspectives:
\textbf{Quantization Efficiency: }We tested the quantization speed of FLRQ against other PTQ methods, with the results presented in Table \ref{quant_time_table}. In the 4-bit and 3-bit quantization, where accuracy drop is less, we selected GPTQ and LQER algorithms for comparison due to faster quantization speed. However, in the 2-bit scenario, AWQ and LQER cannot maintain accuracy; therefore, we chose the higher-precision PTQ methods OmniQuant and AffineQuant for evaluation. The result shows that, in 3 and 4-bit, FLRQ demonstrated comparable quantization times to methods like AWQ, while being over \textbf{30\%} faster than LQER, which employs a SVD. Particularly noteworthy is the performance of FLRQ in the challenging 2-bit quantization scenario, where FLRQ exhibited at least a \textbf{30\%} faster quantization speed compared to OmniQuant, and more than \textbf{5} times faster than AffineQuant.

\begin{figure}[t]
\centering
\includegraphics[width=1\linewidth]{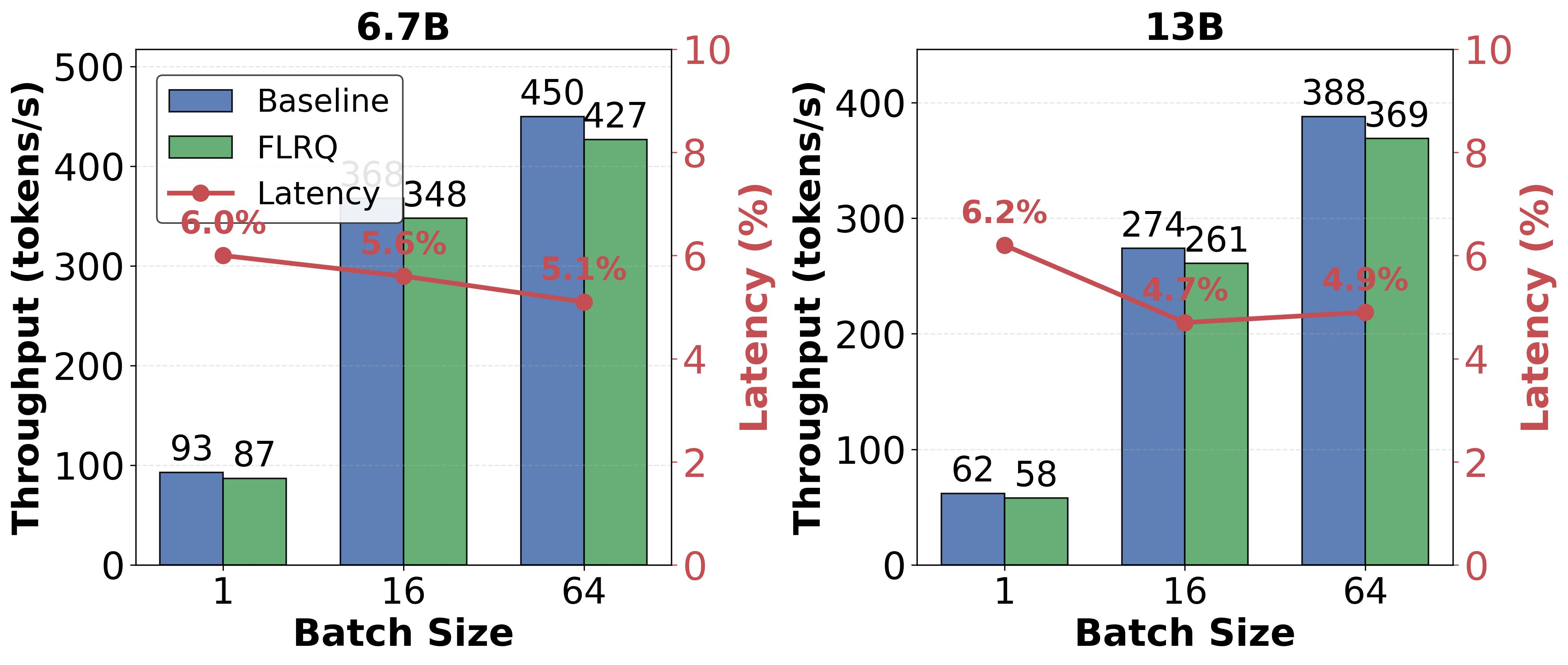}
\caption{ Comparison of Throughput and Latency between Baseline (W4A16) and FLRQ (W4A16+Lora).}
\label{fig_latency}
\end{figure}

\textbf{Inference Efficiency: }
We implemented an efficient fusion kernel for low-rank quantization to the AutoGPTQ quantization framework, with the inference results presented in Figure \ref{fig_latency}. Due to the adoption of the flexible rank selection strategy, FLRQ introduces only a \textbf{4\%} to \textbf{6\%} in inference latency.

\subsection{Ablation studies}
To validate the effectiveness of each strategy within FLRQ, we conducted ablation studies divided into two parts: 

\begin{itemize}
    \item Ablation on \textbf{R1-FLR}: \label{appendix_fix_rank}
To demonstrate the effectiveness of \textbf{R1-FLR}, we compared it with fixed-rank methods without applying calibration. As shown in Table \ref{tab_ab_flr}, FLRQ achieves similar or better PPL to fixed-rank methods, yet with a reduced average bit width, indicating higher compression efficiency. Specifically, in the LLaMA2-13B model, FLRQ achieved the same PPL as a fixed rank of 64 while reducing \textbf{40\%} extra memory costs.

    \item Ablation on \textbf{BLC}: We compare the performance with and without \textbf{BLC}. The results are presented in Table \ref{tab_blc}. And the findings indicate that the use of \textbf{BLC} further reduces the residual space in low-rank quantization, thereby enhancing the accuracy of low-rank quantization at every bit level. Specifically, in the 2-bit scenario, the \textbf{BLC} technique notably avoids quantization distortion, demonstrating superior performance. 
\end{itemize}

{\setlength{\tabcolsep}{4.5pt}
\begin{table}[h]
  \centering

    \small
    \begin{tabular}{ccccccc}
    \Xhline{4\arrayrulewidth}
    \multirow{2}[4]{*}{Bit} & \multirow{2}[4]{*}{Method} & \multicolumn{3}{c}{OPT} & \multicolumn{2}{c}{LLaMA2} \\
\cmidrule{3-7}          &       & 1.3b  & 6.7b  & 13b   & 7b    & 13b \\
    \Xhline{3\arrayrulewidth}
    \multirow{3}[2]{*}{3,4} & AWQ  & 6.7m  & 28.3m & 52.8m & 25.2m & 40.5m \\
          & LQER & 8.2m  & 35.3m & 1.1h  & 45.2m & 1.2h \\
          \rowcolor{black!10} & FLRQ  & 6.2m  & 26.6m & 51.3m & 23.0m & 40.3m\\
    \hline
    \multirow{3}[2]{*}{2} & Omni & 1.2h  & 3.9h  & 6.8h  & 3.1h  & 5.3h \\
          & Affine & 3.2h  & 14.4h & 25.3h & 12.2h & 23.1h \\
          \rowcolor{black!10} & FLRQ  & 33.2m & 2.5h & 4.9h  & 2.0h  & 3.8h \\
    \Xhline{4\arrayrulewidth}
    \end{tabular}%
  \caption{The quantization time costs on a single A100 GPU, where `m' stands for minutes and `h' for hours:}
  \label{quant_time_table}%
\end{table}}

\begin{table}[h]
  \centering

  \setlength{\tabcolsep}{4pt} 
  \small
    \renewcommand{\arraystretch}{1.3}
    \begin{tabular}{ccccccc}
    \Xhline{4\arrayrulewidth}
    \multirow{2}[4]{*}{Models} & \multicolumn{2}{c}{RANK=32} & \multicolumn{2}{c}{RANK=64} & \multicolumn{2}{c}{FLRQ (NO BLC)} \\
\cmidrule{2-7}          & avg. bit & PPL   & avg. bit & PPL   & avg. rank(bit) & PPL \\
    \Xhline{3\arrayrulewidth}
    7B    & 4.32  & 5.83  & 4.52  & 5.73  & 36.03(4.31)$\downarrow$& 5.74 \\
    13B   & 4.28  & 4.99  & 4.44  & 4.98  & 21.9(4.24)$\downarrow$& 4.98 \\
    \Xhline{4\arrayrulewidth}
    \end{tabular}%
  \caption{4-bit PPL under fixed rank and FLRQ on Wiki2.}
  \label{tab_ab_flr}%
\end{table}%
The results indicate that \textbf{R1-FLR} is capable of significantly reducing the additional overhead associated with low-rank quantization while maintaining quantization accuracy and \textbf{BLC} not only improves accuracy across different bit depths but also significantly solves quantization distortion, especially under more challenging 2-bit conditions.
{\setlength{\tabcolsep}{5.5pt}
\begin{table}[h]
  \centering
  \small

    \renewcommand{\arraystretch}{1.2}
    \begin{tabular}{ccccccc}
    
    \Xhline{4\arrayrulewidth}
    \multirow{2}[4]{*}{Bit} & \multirow{2}[4]{*}{BLC} & \multicolumn{3}{c}{OPT} & \multicolumn{2}{c}{LLaMA2} \\
\cmidrule{3-7}          &       & 1.3b  & 6.7b  & 13b   & 7b    & 13b \\
    \Xhline{3\arrayrulewidth}
    \multirow{2}[2]{*}{4} & \texttimes    & 14.58  & 10.89  & 10.11  & 5.55  & 4.94 \\
          & \textbf{\checkmark}   & 14.55$\downarrow$  & 10.84$\downarrow$ & 10.13  & 5.55  & 4.94 \\
    \hline
    \multirow{2}[2]{*}{3} & \texttimes    & 15.80  & 11.32  & 10.54  & 5.89  & 5.18 \\
          & \textbf{\checkmark}   & 15.53$\downarrow$ & 11.18$\downarrow$ & 10.52$\downarrow$ & 5.88$\downarrow$ & 5.16$\downarrow$\\
    \hline
    \multirow{2}[2]{*}{2} & \texttimes     & 29.32  & 17.23  & 15.41  & 2.1e6 & 1.2e6 \\
          & \textbf{\checkmark}   & 22.99$\downarrow$ & 14.05$\downarrow$ & 12.60$\downarrow$ & 9.14$\downarrow$ & 6.77$\downarrow$\\
    \Xhline{4\arrayrulewidth}
    \end{tabular}%
  \caption{PPL($\downarrow$) on WikiText-2 dataset. Note that ``\texttimes" indicates without employing the iterative strategy from \textbf{BLC}. Other settings remain consistent with the main experiments.}
  \label{tab_blc}%
\end{table}}

\section{Conclusion}
In this work, we propose FLRQ, a low-rank quantization method that employs flexible rank selection based on R1-Sketch and iteratively minimizes quantization errors. Compared to other low-rank PTQ, FLRQ features lower additional memory consumption and faster quantization speeds. Extensive experiments demonstrate that FLRQ achieves better precision than several other PTQ methods in model quantization, particularly excelling in 2-bit quantization. Furthermore, we demonstrate the robustness of the FLRQ algorithm and achieve low inference latency through efficient kernel fusion.


\clearpage

\clearpage
\section{Acknowledgements}
This work is partially supported by the Strategic Priority Research Program of Chinese Academy of Sciences (XDB0500101), and the Basic Research Project of the Institute of Software, Chinese Academy of Sciences (ISCAS-JCMS-202304).

\bibliography{aaai2026}

\appendix

\onecolumn

\textbf{Overall:} Comparison between Traditional Low-Rank Quantization and FLRQ. Typically, low-rank PTQs directly apply scaling and low-rank approximation by SVD to the weight matrices. In contrast, FLRQ utilizes flexible rank selections, activation-based scaling,  and iterative Best Low-rank Approximation under Clipping (BLC) algorithm, leading to higher quantization accuracy. Furthermore, R1-Sketch empowers FLRQ to boost efficiency over traditional SVD methods.

\textbf{Organization:} In this appendix, we provide further details as follows:
\begin{itemize}
    \item See Sec.\ref{r1-sketch-appendix}: Presents the details on
    \begin{itemize}
        \item Analysis of Time Complexity \ref{r1-sketch_time}
        \item Necessity of R1-Sketch in FLRQ \ref{Necessity_r1sketch}
        \item The choice of optimal rank in R1-FLR \ref{appendix_mse_rank}
    \end{itemize}      
    \item See Sec.\ref{main_exps_appendix}: Presents the detail experiments results of FLRQ on
    \begin{itemize}
        \item Zero-shot tasks \ref{appendix_zero_shot}
        \item Scaling law \ref{appendix_scaling_law}
        \item Apply R1-Sketch in LQER \ref{r1sketch_in_lqer_appendix}
    \end{itemize}

    \item See Sec.\ref{appendix_hyperarameters}: Presents detailed information on 
    \begin{itemize}
        \item Hyperparameters of memory threshold $x$ \ref{appendix_x}
        \item The number of iterations in R1-Sketch $it$ \ref{appendix_iter}
        \item The epoch of BLC \ref{appendix_epoch}
    \end{itemize}

\clearpage
\end{itemize}

\section{Detail of FLRQ}
\label{r1-sketch-appendix}
\subsection{Analysis of Time Complexity}
\label{r1-sketch_time}
From RSVD, we have R1-Sketch presented as:
\begin{equation}
\begin{split}
  A_L = \boldsymbol Q\boldsymbol  U\boldsymbol  \Sigma &= \frac{\boldsymbol Y}{\Vert \boldsymbol Y\Vert}*\{1\}*\Vert \boldsymbol B \Vert =\frac{(\boldsymbol A\boldsymbol A^*)^{it}\boldsymbol A\boldsymbol S}{\Vert(\boldsymbol A\boldsymbol A^*)^{it}\boldsymbol A\boldsymbol S\Vert}*\frac{\Vert \boldsymbol S^*\boldsymbol A^*(\boldsymbol A\boldsymbol A^*)^{it}\boldsymbol A \Vert}{\Vert(\boldsymbol A\boldsymbol A^*)^{it}\boldsymbol A\boldsymbol S\Vert}\\
  A_R &= \boldsymbol V = \frac{\boldsymbol B}{\Vert \boldsymbol  B\Vert}=\frac{ \boldsymbol S^*\boldsymbol A^*(\boldsymbol A\boldsymbol A^*)^{it}\boldsymbol A }{\Vert \boldsymbol S^*\boldsymbol A^*(\boldsymbol A\boldsymbol A^*)^{it}\boldsymbol A \Vert}.
\end{split}
\label{Lowrank_ALAR}
\end{equation}
Take $\boldsymbol P = (\boldsymbol A\boldsymbol A^*)^{it}\boldsymbol A\boldsymbol S$ and $\boldsymbol K = \boldsymbol A^*\boldsymbol P$. Therefore,
\begin{equation}
\begin{split}
  A_L = \frac{\Vert \boldsymbol K \Vert}{\Vert\boldsymbol P\Vert} \cdot \frac{\boldsymbol P}{\Vert \boldsymbol P\Vert} \in \mathbb{R}^{m \times 1}, \ \ A_R = \frac{\boldsymbol K}{\Vert \boldsymbol K\Vert}\in \mathbb{R}^{1 \times n}.
\end{split}
\label{Lowrank_ALAR_PL}
\end{equation}
Then the rank-1 matrix $\boldsymbol A_1 = \boldsymbol A_L \boldsymbol A_R $ corresponds to the matrix spanned by the singular vector associated with the largest singular value of matrix $\boldsymbol A$. The same process can then be applied to the residual part $\boldsymbol A - \boldsymbol A_1$ to obtain the rank-1 matrix corresponding to the next largest singular value. By iterating this process, we can successively construct a rank-r approximation of $\boldsymbol A$.

Now we conclude that a rank-1 sketch approximation (R1-Sketch) and have Algorithm.\ref{algorithmofbestrank}. Since R1-Sketch is entirely derived from the RSVD algorithm under the rank-1 condition, it maintains the same accuracy and error bounds as the RSVD algorithm.

While the norms of $\Vert \boldsymbol K\Vert$ and $\Vert \boldsymbol P\Vert$ can be efficiently computed with a time complexity of $O(n)$. Consequently, the main computational cost of the algorithm is the calculation of $\boldsymbol P$ and $\boldsymbol K$, which is composed entirely of GEMV (matrix-vector multiplication), and with a time complexity of $O((2{it}+2)n^2)$. As the parameter $it$ increases, the algorithm achieves higher accuracy at the cost of greater computational overhead. However, in practical applications, $it$ does not need to be very large (approximately 2) to achieve good approximation results in the context of llm quantization. We will provide a further discussion Section \ref{appendix_iter}.

\begin{algorithm}[H]
   \caption{R1-Sketch-based Flexible Low-Rank Selection (\textbf{R1-FLR})}
   \label{algorithmofbestrank}
   \KwData {$\boldsymbol W$(weight matrix), $d$(bits), $x$(maximum model size increase)}
   \KwResult{$\boldsymbol W_L, \boldsymbol W_R$}
    get shape and maximum rank: $\{m,n\} \leftarrow \boldsymbol W.shape, R \leftarrow min(m,n)$\;
    get origin maximum absolute value: $maxAbs_0 \leftarrow max(abs(\boldsymbol W))$\;
    \For{$i=1$ {\bfseries to} $R$}{
     Obtain the sketch rank-1 matrix corresponding to the largest singular value: $\{U_1, V_1\} \leftarrow calR1matrix(\boldsymbol W)$ \;
     $\boldsymbol W \leftarrow  \boldsymbol W - \boldsymbol U_1\cdot \boldsymbol V_1$\;
     get maximum absolute value now: $maxAbs \leftarrow  max(abs(\boldsymbol W))$\;
     Calculate P, Q, K: $P \leftarrow  maxAbs_0\leftarrow maxAbs$\;
     $Q \leftarrow (d + log_2(P) )/d)$\;
     $K \leftarrow 1 + (d*r*(m+n))/(m*n))$\;
     Calculate the slope of maxAbs: $sNow \leftarrow  getSlope(maxAbs)$\;
    \If{$K > Q \ or \ K > 1+x \ or\ sNow < t$}{
         Endloop\;
    }
     $\boldsymbol W_L\leftarrow \boldsymbol W_L$.append($U_1$)\;
     $\boldsymbol W_R\leftarrow \boldsymbol W_R$.append($V_1$)\;
    }
   \Return{ $\boldsymbol W_L, \boldsymbol W_R$}
\end{algorithm}

\begin{algorithm}[H]
   \caption{R1-Sketch-based Flexible Low-Rank Selection (\textbf{R1-FLR})}
   \label{algorithmofbestrank}
   \KwData {$\boldsymbol W$(weight matrix), $d$(bits), $x$(maximum model size increase)}
   \KwResult{$\boldsymbol W_L, \boldsymbol W_R$}
    \For{$i=1$ {\bfseries to} $R$}{
     $\{U_1, V_1\} \leftarrow calR1matrix(\boldsymbol W)$ \;
     $\boldsymbol W \leftarrow  \boldsymbol W - \boldsymbol U_1\cdot \boldsymbol V_1$\;
    }
   \Return{ $\boldsymbol W_L, \boldsymbol W_R$}
\end{algorithm}

\subsection{Discussion on the necessity of R1-Sketch}
\label{Necessity_r1sketch}
Why is R1-Sketch necessary? Given that the algorithm is derived from RSVD, and its computational accuracy is the same as that of the RSVD algorithm, the necessity of R1-Sketch mainly lies in considerations of computational efficiency. The sole difference between R1-Sketch and RSVD or truncated SVD is that the construction of the sketch matrix in R1-Sketch is discrete; that is, for each rank, the determination of $absmax$ or error $\mathbb{E}$ can be achieved through a single rank-1 approximation. If the condition for optimal rank is met, the computation stops immediately without any redundant calculations.

However, if the SVD algorithm is applied, it requires decomposing the entire matrix. Similarly, using RSVD or truncated SVD requires to specify the rank $r$ of the approximate matrix beforehand. But before calculating the error or absmax value, the optimal rank value cannot be known, making it necessary to approximate with a relatively large rank across all layers and then traverse every sub-matrix of this rank-r matrix to determine the best rank. Sub-matrices beyond the best rank are essentially unused, representing redundant computations. In the case of the LLaMA2-7b model, the optimal rank sizes for different layers are shown in Table \ref{best_rank_static}, varying from 0 to 128. Using RSVD would require approximating with a larger rank value (above 128) followed by the application of FLR techniques. Due to BLC technology, FLRQ under low-bit conditions requires more iterations to reduce the residual space, necessitating multiple low-rank approximation functions. Decomposing with a large rank consumes considerable computing resources unnecessarily.
\begin{table}[H]
  \centering
  \caption{In the LLaMA2-7b model, the statistical results of the best rank across different layers.}
    \begin{tabular}{c|ccccccc}
    \toprule
    Rank  & 0$ \sim $8   & 8$ \sim $16  & 16$ \sim $32 & 32$ \sim $48 & 48$ \sim $64 & 64$ \sim $128 & avg.rank \\
    \midrule
    Layer-Nums & 7     & 19    & 44    & 61    & 37    & 5     & 35.29 \\
    \bottomrule
    \end{tabular}%
  \label{best_rank_static}%
\end{table}%
For truncated SVD, we set the truncation rank to 128 for models up to 7 billion parameters, and to 256 for the 13 billion parameter model (this setting is due to the best rank for certain layers in the 13 billion model reaching 237). We measured the quantization time, and the results are reported in Table \ref{tsvd_time}. The results show that the computation time using truncated SVD is significantly high, whereas R1-Sketch effectively reduces the computational load, greatly enhancing computational efficiency. This also demonstrates the effectiveness and necessity of R1-Sketch within the FLRQ algorithm.

\begin{table}[H]
  \centering
  \caption{The runtime for model quantization when using Truncated SVD (T-SVD) and R1-Sketch in the FLRQ on different models.}
    \begin{tabular}{cccccccc}
    \toprule
    \multirow{2}[4]{*}{Precision} & \multirow{2}[4]{*}{Method} & \multicolumn{4}{c}{OPT}       & \multicolumn{2}{c}{LLaMA2} \\
\cmidrule{3-8}          &       & 1.3b  & 2.7b  & 6.7b  & 13b   & 7b    & 13b \\
    \midrule
    \multirow{2}[2]{*}{3bit,4bit} & FLRQ(T-SVD) & 16.0m & 39.5m & 1.4h  & 3.3h  & 1.1h  & 3.1h \\
          & FLRQ(R1-Sketch) & 6.2m  & 12.3m & 26.6m & 51.3m & 23.0m & 40.3m \\
    \midrule
    \multirow{2}[2]{*}{2bit} & FLRQ(T-SVD) & 1.3h  & 2.1h  & 5.9h  & 12.8h & 5.2h  & 11.3h \\
          & FLRQ(R1-Sketch) & 33.2m & 1.1h  & 2.5h  & 4.9h  & 2.0h  & 3.8h \\
    \bottomrule
    \end{tabular}%
  \label{tsvd_time}%
\end{table}%

\subsection{The choice of optimal rank in R1-FLR}
\label{appendix_mse_rank}
\begin{figure}[H]
\begin{center}
\centerline{\includegraphics[width=1\linewidth]{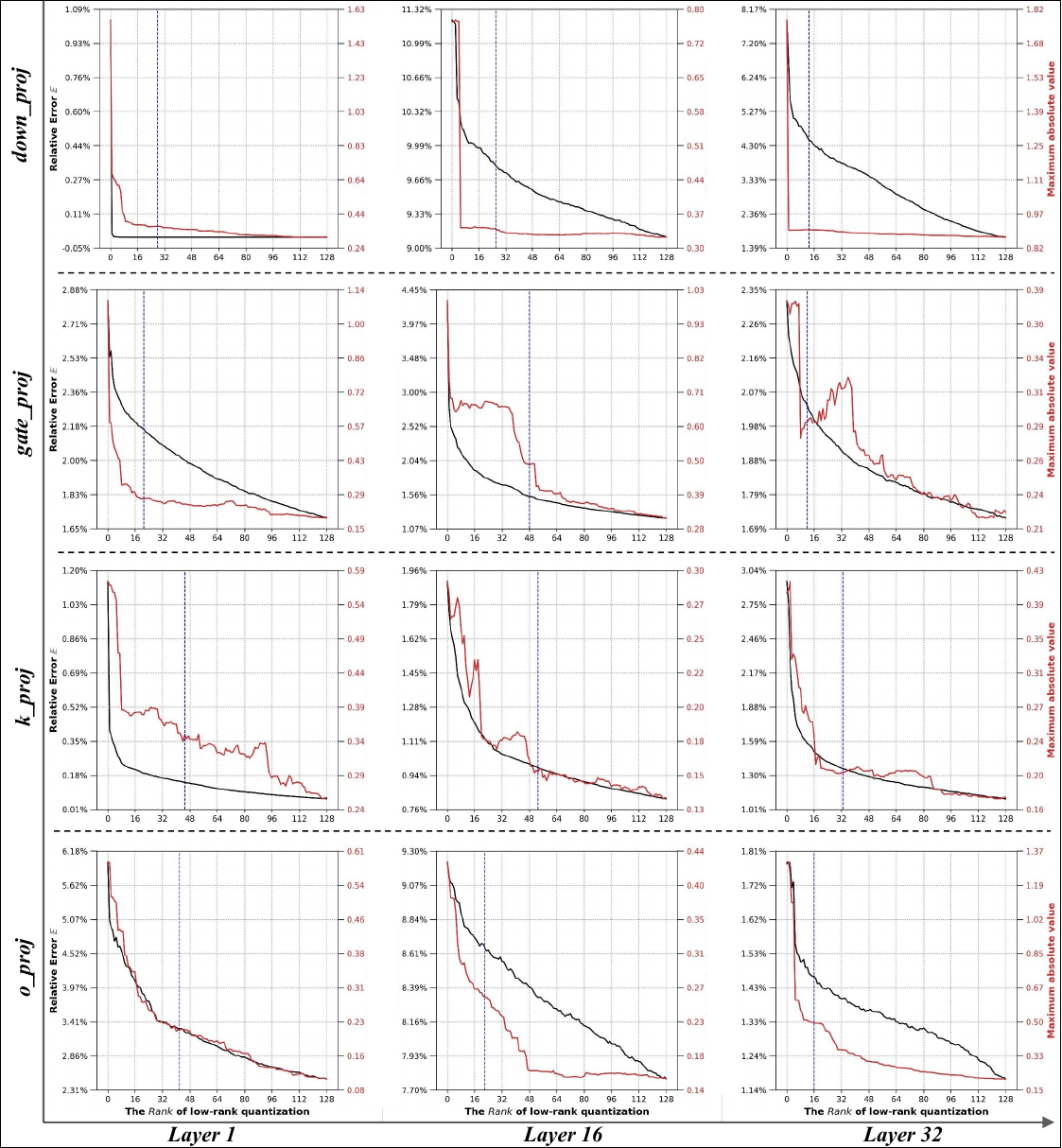}}
\caption{More visualization in LLaMA2-7b on the relationship between rank selection and the error $\mathbb{E}$/$absmax$, where the blue lines (\textcolor{blue}{- - -}) stand for the optimal rank from R1-FLR. Results demonstrate how varying the rank affects the quantization error, providing insights into the trade-offs between different rank choices and their corresponding impact on quantization accuracy. }
\label{llama_mse_max_fig}
\end{center}
\vskip -0.2in
\end{figure}

We present the variation curves of low-rank quantization error and $absmax$ with increasing rank in the LLaMA2-7b model in Figure \ref{llama_mse_max_fig}. Across different layers, the error curves are generally smooth but exhibit varying rates of decrease. The $absmax$ variation curves closely resemble those of the error E. In most cases, R1-FLR using $absmax$ can effectively determine an optimal rank. For instance, in \textit{Layer32-down}, using a rank of just 15 achieves nearly a 50\% reduction in error, while in \textit{Layer1-k}, a rank of 44 results in an 85\% reduction in error.

However, due to the quantization strategy involving repeated iterations and subsequent computations, the $absmax$ curve does not always perfectly align with the final error curve. For example, in \textit{Layer1-down}, the error is nearly zero at a rank of around 2, with minimal changes thereafter, whereas the $absmax$ curve continues to show significant decreases around this rank, leading to an estimated optimal rank that is disproportionately high, around 30. Although there are instances where the error curve and $absmax$ curve do not fit well, the flexible rank selection approach still better accommodates the characteristics of different layers compared to using a fixed rank.

\section{Supplementary EXPERIMENT DETAILS}
\label{main_exps_appendix}
\subsection{ZERO-SHOT TASKS}
\label{appendix_zero_shot}
This section contains additional results for zero-shot tasks. Here \textbf{OpenbookQA} is in short of \textbf{OB-QA}, and \textbf{ARC-C}, \textbf{ARC-E} stands for \textbf{ARC-challenge} and \textbf{ARC-easy}. The introduction of the datasets are as follows:
\begin{itemize}
    \item ARC-Challenge: This dataset contains science exam questions (typically multiple-choice) specifically curated to be difficult for retrieval-based and word co-occurrence methods.
    \item ARC-Easy: This counterpart contains science questions from the same source (elementary and middle school level) that are more readily answerable by simple information retrieval or shallow methods. 
    \item BoolQ (Boolean Questions): This dataset consists of yes/no questions based on short passages of text. The task requires a model to read a passage and a related question, then determine the correct boolean answer (True/False or Yes/No). 
    \item OpenBookQA: This dataset is built around a "book" of elementary-level science facts. Each question requires the model to use one or more of these core facts in conjunction with common knowledge to perform multi-step reasoning and answer the question correctly.
    \item PIQA: This dataset focuses on commonsense physical reasoning. It presents questions about the physical world (e.g., "How would one go about hanging a picture?") with two possible answer choices. The goal is to select the more physically plausible or sensible solution. 
    \item Winogrande: This dataset is designed to evaluate commonsense reasoning, specifically coreference resolution, using a Winograd Schema format. It presents sentences with ambiguous pronouns, where resolving the correct referent requires understanding the context and applying real-world knowledge.
\end{itemize}

\begin{table}[H]
  \centering
  \caption{OPT-1.3b}
  \label{OPT-1.3b-zero}%
    \begin{tabular}{ccccccccc}
    \toprule
    \multicolumn{2}{c}{OPT-1.3b} & ARC-C & ARC-E & BOOLQ & OB-QA & PIQA  & Wino & Avg. Acc \\
    \midrule
    \multicolumn{2}{c}{FP16} & 23.29\% & 57.03\% & 57.83\% & 23.40\% & 71.60\% & 59.51\% & 48.78\% \\
    \hdashline
    \multirow{3}[2]{*}{4bit} & Omniquant & 23.41\% & 55.33\% & 56.21\% & 22.22\% & 71.13\% & 58.46\% & 47.79\% \\
          & AWQ   & 23.31\% & 53.34\% & 55.15\% & 22.40\% & 71.81\% & 58.10\% & 47.35\% \\
          & FLRQ  & 23.81\% & 57.28\% & 57.68\% & 22.00\% & 71.06\% & 58.64\% & 48.41\% \\
    \midrule
    \multirow{3}[2]{*}{3bit} & Omniquant & 23.13\% & 54.97\% & 56.76\% & 21.32\% & 70.51\% & 57.51\% & 47.37\% \\
          & AWQ   & 21.01\% & 54.63\% & 54.16\% & 22.00\% & 70.78\% & 56.88\% & 46.58\% \\
          & FLRQ  & 22.35\% & 55.05\% & 59.82\% & 20.60\% & 70.08\% & 57.77\% & 47.61\% \\
    \midrule
    \multirow{2}[2]{*}{2bit} & Omniquant & 19.98\% & 45.15\% & 57.76\% & 16.62\% & 60.79\% & 51.33\% & 41.94\% \\
          & FLRQ  & 22.53\% & 48.86\% & 61.28\% & 17.60\% & 66.49\% & 56.51\% & 45.55\% \\
    \bottomrule
    \end{tabular}%
  \label{tab:addlabel}%
\end{table}%

\begin{table}[H]
  \centering
  \caption{OPT-6.7B}
    \begin{tabular}{ccccccccc}
    \toprule
    \multicolumn{2}{c}{OPT-6.7b} & ARC-C & ARC-E & BOOLQ & OB-QA & PIQA  & Wino  & Avg. Acc \\
    \midrule
    \multicolumn{2}{c}{FP16} & 30.46\% & 65.61\% & 66.09\% & 27.60\% & 76.28\% & 65.19\% & 55.21\% \\
    \hdashline
    \multirow{3}[2]{*}{4bit} & Omniquant & 30.62\% & 65.61\% & 66.21\% & 26.77\% & 75.98\% & 64.32\% & 54.92\% \\
          & AWQ   & 30.50\% & 65.30\% & 65.20\% & 26.60\% & 76.60\% & 64.20\% & 54.73\% \\
          & FLRQ  & 29.78\% & 65.78\% & 66.64\% & 27.20\% & 75.95\% & 64.88\% & 55.04\% \\
    \midrule
    \multirow{3}[2]{*}{3bit} & Omniquant & 29.33\% & 64.13\% & 65.12\% & 24.98\% & 74.56\% & 64.04\% & 53.69\% \\
          & AWQ   & 28.98\% & 63.76\% & 64.77\% & 24.20\% & 74.24\% & 63.35\% & 53.22\% \\
          & FLRQ  & 29.86\% & 64.27\% & 65.60\% & 25.80\% & 75.73\% & 64.96\% & 54.37\% \\
    \midrule
    \multirow{2}[2]{*}{2bit} & Omniquant & 22.44\% & 56.76\% & 56.42\% & 21.53\% & 66.77\% & 57.56\% & 46.91\% \\
          & FLRQ  & 26.54\% & 60.35\% & 63.33\% & 24.80\% & 72.69\% & 61.01\% & 51.45\% \\
    \bottomrule
    \end{tabular}%
  \label{OPT-6.7b-zero}%
\end{table}%

\begin{table}[H]
  \centering
  \caption{OPT-13B}
    \begin{tabular}{ccccccccc}
    \toprule
    \multicolumn{2}{c}{OPT-13b} & ARC-C & ARC-E & BOOLQ & OB-QA & PIQA  & Wino  & Avg. Acc \\
    \midrule
    \multicolumn{2}{c}{FP16} & 32.85\% & 67.09\% & 65.87\% & 27.00\% & 75.84\% & 65.19\% & 55.64\% \\
    \hdashline
    \multirow{3}[2]{*}{4bit} & Omniquant & 33.51\% & 66.78\% & 66.02\% & 26.77\% & 75.47\% & 65.25\% & 55.63\% \\
          & AWQ   & 33.20\% & 66.80\% & 66.50\% & 28.00\% & 75.60\% & 64.30\% & 55.73\% \\
          & FLRQ  & 34.04\% & 66.92\% & 68.65\% & 26.80\% & 75.52\% & 65.59\% & 56.25\% \\
    \midrule
    \multirow{3}[2]{*}{3bit} & Omniquant & 30.87\% & 65.66\% & 68.42\% & 26.13\% & 75.83\% & 61.82\% & 54.79\% \\
          & AWQ   & 30.12\% & 63.24\% & 68.12\% & 24.14\% & 73.24\% & 61.88\% & 53.46\% \\
          & FLRQ  & 30.89\% & 65.36\% & 69.57\% & 26.20\% & 76.06\% & 63.77\% & 55.31\% \\
    \midrule
    \multirow{2}[2]{*}{2bit} & Omniquant & 20.44\% & 54.21\% & 58.32\% & 22.13\% & 69.95\% & 58.89\% & 47.32\% \\
          & FLRQ  & 27.73\% & 61.24\% & 64.53\% & 24.20\% & 72.80\% & 63.30\% & 52.30\% \\
    \bottomrule
    \end{tabular}%
  \label{OPT-13B-zero}%
\end{table}%

\begin{table}[H]
  \centering
  \caption{LLaMA2-7B}
    \begin{tabular}{ccccccccc}
    \toprule
    \multicolumn{2}{c}{LLaMA-2-7b} & ARC-C & ARC-E & BOOLQ & OB-QA & PIQA  & Wino  & Avg. Acc \\
    \midrule
    \multicolumn{2}{c}{FP16} & 43.43\% & 76.35\% & 77.71\% & 31.40\% & 78.07\% & 69.22\% & 62.70\% \\
    \hdashline
    \multirow{3}[2]{*}{4bit} & Omniquant & 43.32\% & 76.01\% & 77.75\% & 31.50\% & 77.67\% & 68.89\% & 62.52\% \\
          & AWQ   & 43.30\% & 75.20\% & 77.30\% & 31.40\% & 77.60\% & 68.20\% & 62.17\% \\
          & FLRQ  & 43.77\% & 76.39\% & 77.61\% & 31.60\% & 77.58\% & 69.14\% & 62.68\% \\
    \midrule
    \multirow{3}[2]{*}{3bit} & Omniquant & 40.78\% & 74.49\% & 74.28\% & 31.88\% & 77.75\% & 67.64\% & 61.14\% \\
          & AWQ   & 39.22\% & 73.88\% & 74.87\% & 31.35\% & 77.24\% & 68.98\% & 60.92\% \\
          & FLRQ  & 40.36\% & 73.95\% & 75.08\% & 32.00\% & 77.53\% & 69.30\% & 61.37\% \\
    \midrule
    \multirow{2}[2]{*}{2bit} & Omniquant & 28.84\% & 58.12\% & 60.33\% & 24.14\% & 70.17\% & 59.12\% & 50.12\% \\
          & FLRQ  & 35.49\% & 67.72\% & 70.09\% & 26.20\% & 74.48\% & 64.64\% & 56.44\% \\
    \bottomrule
    \end{tabular}%
  \label{LLaMA-2-7B-zero}%
\end{table}%

\begin{table}[H]
  \centering
  \caption{LLaMA2-13B}
    \begin{tabular}{ccccccccc}
    \toprule
    \multicolumn{2}{c}{LLaMA-2-13b} & ARC-C & ARC-E & BOOLQ & OB-QA & PIQA  & Wino  & Avg. Acc \\
    \midrule
    \multicolumn{2}{c}{FP16} & 49.10\% & 77.44\% & 80.71\% & 35.90\% & 81.35\% & 73.89\% & 66.40\% \\
    \hdashline
    \multirow{3}[2]{*}{4bit} & Omniquant & 47.63\% & 78.82\% & 79.13\% & 33.35\% & 79.01\% & 71.89\% & 64.97\% \\
          & AWQ   & 46.90\% & 78.90\% & 77.30\% & 31.40\% & 77.60\% & 72.40\% & 64.08\% \\
          & FLRQ  & 48.21\% & 79.97\% & 79.45\% & 34.40\% & 78.56\% & 71.82\% & 65.40\% \\
    \midrule
    \multirow{3}[2]{*}{3bit} & Omniquant & 42.00\% & 77.94\% & 79.03\% & 32.25\% & 77.98\% & 71.32\% & 63.42\% \\
          & AWQ   & 42.04\% & 77.33\% & 78.22\% & 31.85\% & 76.14\% & 70.53\% & 62.69\% \\
          & FLRQ  & 45.22\% & 78.37\% & 79.08\% & 32.80\% & 78.40\% & 71.27\% & 64.19\% \\
    \midrule
    \multirow{2}[2]{*}{2bit} & Omniquant & 31.33\% & 62.31\% & 64.09\% & 27.52\% & 72.33\% & 65.65\% & 53.87\% \\
          & FLRQ  & 39.16\% & 72.05\% & 76.45\% & 30.20\% & 76.12\% & 68.35\% & 60.39\% \\
    \bottomrule
    \end{tabular}%
  \label{LLaMA-2-13B-zero}%
\end{table}%

\subsection{Scaling laws}
\label{appendix_scaling_law}
\begin{figure}[H]
\begin{center}
\centerline{\includegraphics[width=1\linewidth]{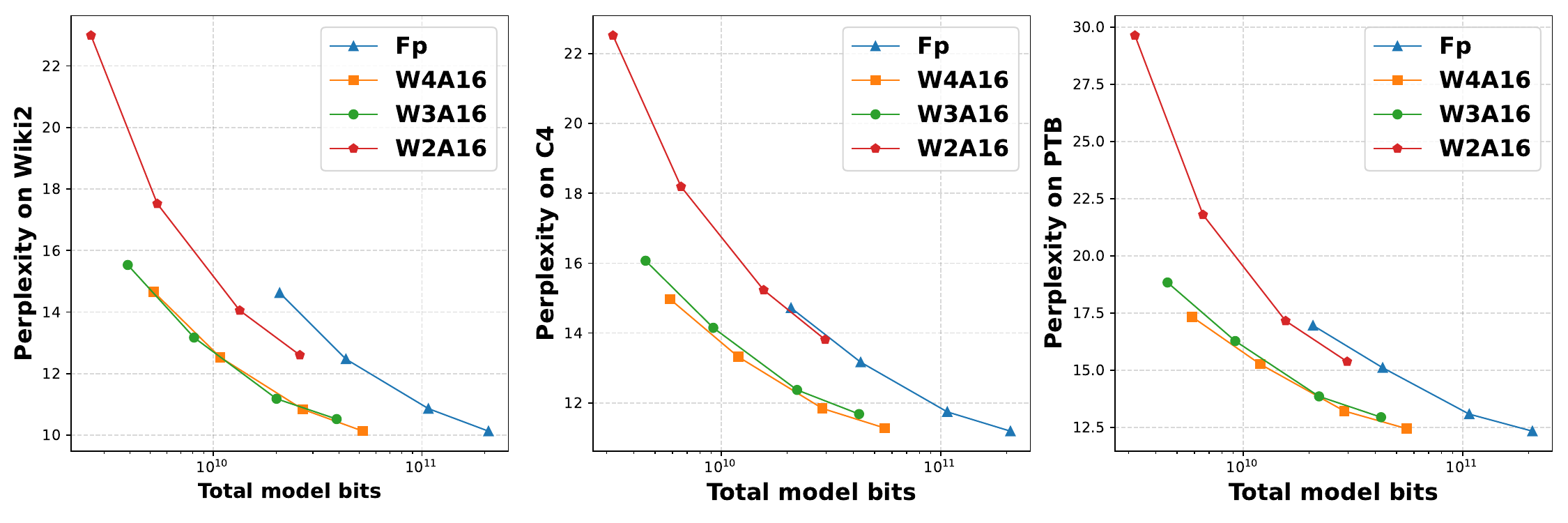}}
\caption{The scaling laws of perplexity under bit-level.}
\label{scaling_law_fig}
\end{center}
\vskip -0.2in
\end{figure}

Quantization is an effective strategy to reduce the total model size, thereby facilitating the deployment of large language models (LLMs) on memory-constrained edge or consumer devices. An excellent quantization algorithm should maintain effective accuracy across a range of models from small to large, meaning that the quantization effectiveness should improve in terms of accuracy as the model size increases across different bit widths. We present the perplexity (PPL) results of Fixed Low-Rank Quantization (FLRQ) across various datasets in Figure \ref{scaling_law_fig}. The results demonstrate that FLRQ maintains sufficiently impressive scaling effects, particularly achieving comparable performance to 4-bit quantization in terms of the trade-off between model size and PPL with its 3-bit quantization.

\subsection{Apply R1-Sketch in LQER}
\label{r1sketch_in_lqer_appendix}
In this section, we present experiments comparing the original low-rank approximation implementation (\textit{torch.linear.svd}) in the LQER algorithm with its replacement by \textbf{\textit{R1-Sketch}}, evaluating both PPL and execution time in low-rank approximation. The results demonstrate that R1-Sketch significantly improves algorithm execution speed while maintaining lossless performance in terms of PPL.

\begin{table}[H]
\centering
  \caption{Thet WikiText2 PPL performance of L$^2$QER on different models, with quantization parameters set to W4A16, rank $ r = 32$, and group size is 128.}
    \begin{tabular}{lcccc}
    \toprule
    \multicolumn{1}{c}{\multirow{2}[4]{*}{Method}} & \multicolumn{2}{c}{OPT} & \multicolumn{2}{c}{LLaMA-2} \\
\cmidrule{2-5}          & 6.7B  & 13B   & 7B    & 13B \\
    \midrule
    \multicolumn{1}{c}{FP16} & 10.86 & 10.13 & 5.48  & 4.88 \\
    \multicolumn{1}{c}{L$^2$QER-svd} & 10.99 & 10.24 & 5.58  & 4.96 \\
    L$^2$QER-sketch & 10.99 & 10.23 & 5.58  & 4.96 \\
    \bottomrule
    \end{tabular}%
  \label{lqer-sketch-ppl}%
\end{table}

\begin{figure}[H]
\begin{center}
\centerline{\includegraphics[width=0.8\linewidth]{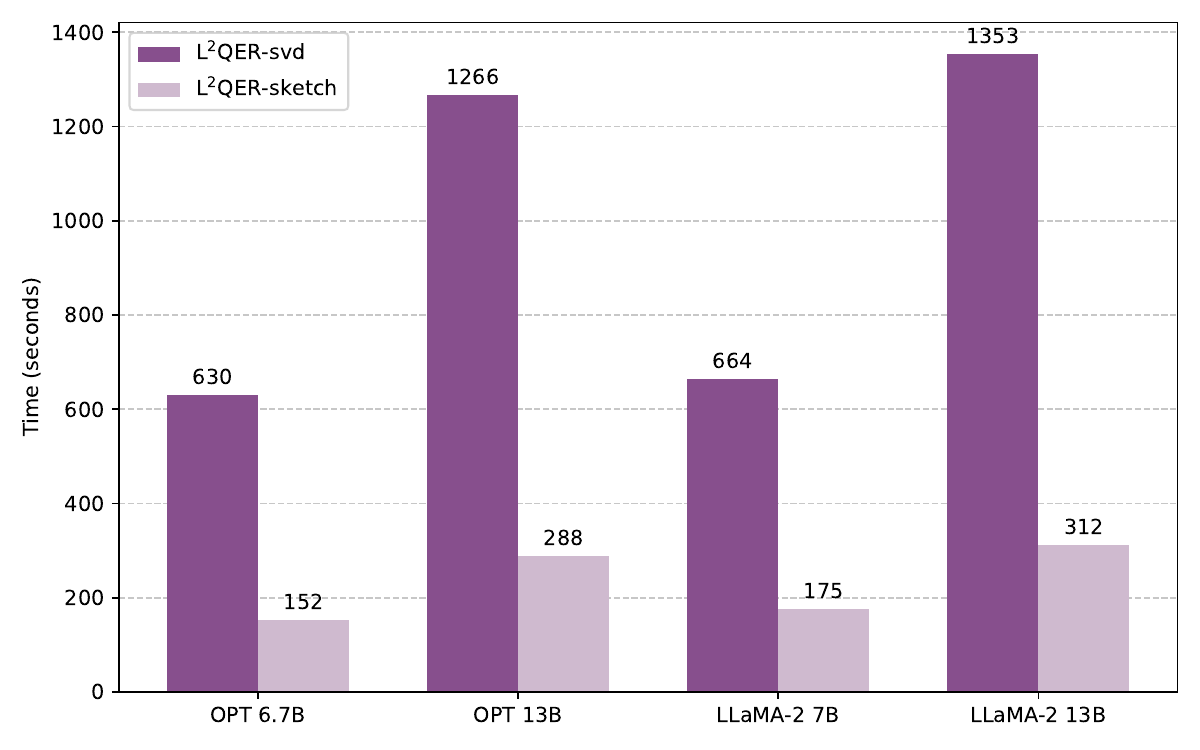}}
\caption{R1-Sketch is an accuracy-preserving low-rank approximation method designed for large-scale model quantization scenarios. Compared to the built-in SVD function in PyTorch, the rank-32 low-rank approximation in L$^2$QER achieves a maximum of $4.4\times$ speedup during the approximation stage.}
\label{lqer_time}
\end{center}
\vskip -0.2in
\end{figure}

\section{Detailed experiments on hyperparameter selection}
\label{appendix_hyperarameters}
\subsection{The setting of memory limit \textit{x}}
\label{appendix_x}
In FLRQ, we introduced a hyperparameter \( x \) to represent our desired maximum increase in the size of the model. In this context, if the memory usage of the low-rank component larger than $x$, it indicates that the additional memory overhead introduced by low-rank components has approached the acceptable maximum size (i.e., the model cannot grow further!). At this point, the iteration process will terminate immediately. 

The optimal rank selection strategy in \textbf{R1-FLR} is not directly related to $x$; the threshold $x$ only determines the maximum value for rank selection. We show the rank / extra average bit width of FLRQ at different $x$ values in Table \ref{x_rank_bit} and actual memory consumption in Table \ref{act_memory} together with Wikitext2-PPL in Table \ref{x_ppl}. 

From the experimental results, it shows that for smaller models (OPT-125M), since the effective rank of the matrix in the quantization scenario is close to the overall size of the matrix, a larger threshold (e.g., 0.4) can lead to higher quantization accuracy but also significantly increases the average number of quantized bits. However, for larger models (OPT-2.7B and more), where the matrix dimensions far exceed the rank of low-rank components extracted by \textbf{R1-FLR}, the performance of FLRQ is similar across different thresholds. 

From another perspective, as the model size increases, the average number of bits extracted by FLRQ gradually decreases. In the 2-bit, 3-bit, and 4-bit quantization scenarios for the 13B model, the low-rank components only occupy around 0.2 bits. For example, under W4A16G128-OPT13B, the total average number of bits for FLRQ is 4(4bit quantization) + 0.12(low-rank component of FLRQ) + 0.16 (for storing scale and zero with a group size of 128)->  \textbf{total = 4.28 bits}. Therefore, FLRQ also exhibits excellent scalability from a memory perspective on larger models.

In terms of actual memory consumption, since the quantization does not quantize every weight matrix, FLRQ's actual memory footprint is less than the average extra bits usage! In the LLaMA2-13B model, at a $x=0.2$, the additional memory usage of FLRQ is only approximately 4\% of that in 3-bit quantization and about 6\% of that in 2-bit quantization.

For the reasons mentioned above, we uniformly set the threshold $x = 0.2$ in other experiments. This setting ensures model accuracy across various scales while preventing excessive additional memory consumption.

\begin{table}[H]
  \centering
  \caption{The extracted rank and extra average bit width of FLRQ at different $x$ values, represented as (rank/extra-avg.bit). The corresponding PPL results can be found in Table \ref{x_ppl}. In other experiments, the x value is set to $0.2$, with these results highlighted in a darker color.}
    \begin{tabular}{ccccccccc}
    \toprule
    \multirow{2}[4]{*}{Bit} & \multirow{2}[4]{*}{x} & \multicolumn{5}{c}{OPT}               & \multicolumn{2}{c}{LLaMA2} \\
\cmidrule{3-9}          &       & 125M  & 1.3B  & 2.7B  & 6.7B  & 13B   & 7B    & 13B \\
    \midrule
    \multirow{3}[2]{*}{4} & 0.1   & 12.3/0.36 & 27.0/0.31 & 25.4/0.26 & 27.0/0.16 & 26.4/0.12 & 34.9/0.20 & 38.2/0.18 \\
          \rowcolor{black!10} 4 & 0.2   & 19.9/0.64 & 30.5/0.34 & 29.6/0.27 & 27.1/0.16 & 27.0/0.12 & 36.1/0.21 & 38.6/0.18 \\
          & 0.4   & 27.7/0.77 & 31.32/0.35 & 32.3/0.28 & 27.8/0.16 & 27.2/0.12 & 36.4/0.21 & 38.7/0.18 \\
    \midrule
    \multirow{3}[2]{*}{3} & 0.1   & 8.3/0.28 & 21.0/0.26 & 24.4/0.23 & 26.6/0.15 & 26.5/0.12 & 35.1/0.20 & 38.0/0.18 \\
          \rowcolor{black!10} 3 & 0.2   & 16.1/0.54 & 28.8/0.33 & 28.9/0.26 & 27.7/0.16 & 26.4/0.12 & 35.8/0.21 & 38.4/0.18 \\
          & 0.4   & 25.1/0.75 & 29.6/0.34 & 30.2/0.27 & 27.8/0.16 & 26.4/0.12 & 36.0/0.21 & 38.5/0.18 \\
    \midrule
    \multirow{3}[2]{*}{2} & 0.1   & 5.6/0.19 & 18.0/0.19 & 18.4/0.18 & 27.0/0.16 & 27.3/0.12 & 36.3/0.22 & 40.3/0.19 \\
          \rowcolor{black!10} 2 & 0.2   & 10.4/0.38 & 27.6/0.33 & 30.9/0.29 & 32.7/0.19 & 33.6/0.15 & 39.2/0.24 & 41.9/0.20 \\
          & 0.4   & 21.2/0.70 & 28.7/0.34 & 32.1/0.30 & 34.1/0.20 & 33.9/0.15 & 40.1/0.25 & 42.1/0.20 \\
    \bottomrule
    \end{tabular}%
  \label{x_rank_bit}%
\end{table}%

\begin{table}[htbp]
  \centering
  \caption{The  memory costs of FLRQ under different values of $x$, where $x = 0$ corresponds to no additional low-rank matrices being used, i.e., the original W(2,3,4)G128 quantization.}
    \begin{tabular}{ccccccccc}
    \toprule
    \multirow{2}[4]{*}{Bit} & \multirow{2}[4]{*}{x} & \multicolumn{5}{c}{OPT}               & \multicolumn{2}{c}{LLaMA2} \\
\cmidrule{3-9}          &       & 125M  & 1.3B  & 2.7B  & 6.7B  & 13B   & 7B    & 13B \\
    \midrule
    \multicolumn{2}{c}{fp16} & 313MB & 2.52GB & 5.21GB & 12.6GB & 25.71GB & 13.48GB & 26.03 \\
    \midrule
    \multirow{4}[1]{*} & 0     & 125MB & 1050MB & 1.84GB & 4.22GB & 7.63GB & 3.89GB & 7.25GB \\
    \hdashline
          & 0.1   & 128MB & 1095MB & 1.91GB & 4.32GB & 7.79GB & 4.07GB & 7.46GB \\
          \rowcolor{black!10}4& 0.2   & 130MB & 1101MB & 1.92GB & 4.32GB & 7.79GB & 4.08GB & 7.47GB \\
          & 0.4   & 132MB & 1102MB & 1.93GB & 4.32GB & 7.79GB & 4.08GB & 7.47GB \\
    \midrule
    \multirow{4}[2]{*} & 0     & 114MB & 859MB & 1.53GB & 3.09GB & 5.73GB & 3.08GB & 5.43GB \\
    \hdashline
          & 0.1   & 116MB & 894MB & 1.60GB & 3.20GB & 5.92GB & 3.22GB & 5.57GB \\
          \rowcolor{black!10}3& 0.2   & 118MB & 907MB & 1.61GB & 3.21GB & 5.92GB & 3.27GB & 5.64GB \\
          & 0.4   & 120MB & 908MB & 1.62GB & 3.21GB & 5.92GB & 3.27GB & 5.64GB \\
    \midrule
    \multirow{4}[2]{*} & 0     & 101MB & 713MB & 1.20GB & 2.30GB & 4.21GB & 2.27GB & 3.93GB \\
    \hdashline
          & 0.1   & 102MB & 743MB & 1.25GB & 2.42GB & 4.4GB & 2.46GB & 4.15GB \\
          \rowcolor{black!10}2& 0.2   & 104MB & 759MB & 1.29GB & 2.44GB & 4.45GB & 2.47GB & 4.16GB \\
          & 0.4   & 106MB & 761MB & 1.29GB & 2.45GB & 4.45GB & 2.48GB & 4.16GB \\
    \bottomrule
    \end{tabular}%
  \label{act_memory}%
\end{table}%

\begin{table}[H]
  \centering
  \caption{The PPL($\downarrow$) of FLRQ on Wikitext2 at different $x$ values. The corresponding rank/extra-avg.bit results can be found in Table \ref{x_rank_bit}. In other experiments, the $x$ value is set to $0.2$, with these results highlighted in a darker color.}
    \begin{tabular}{ccccccccc}
    \toprule
    \multirow{2}[4]{*}{Bit} & \multirow{2}[4]{*}{x} & \multicolumn{5}{c}{OPT}               & \multicolumn{2}{c}{LLaMA2} \\
\cmidrule{3-9}          &       & 125M  & 1.3B  & 2.7B  & 6.7B  & 13B   & 7B    & 13B \\
    \midrule
    \multirow{3}[2]{*}{4} & 0.1   & 30.31  & 14.69  & 12.61  & 10.84  & 10.15  & 5.56  & 4.94  \\
          \rowcolor{black!10} 4 & 0.2   & 28.65  & 14.65  & 12.53  & 10.84  & 10.13  & 5.55  & 4.94  \\
          & 0.4   & 26.98  & 14.65  & 12.52  & 10.84  & 10.13  & 5.55  & 4.94  \\
    \midrule
    \multirow{3}[2]{*}{3} & 0.1   & 34.15  & 15.78  & 13.29  & 11.19  & 10.60  & 5.89  & 5.17  \\
         \rowcolor{black!10} 3 & 0.2   & 32.55  & 15.53  & 13.17  & 11.18  & 10.59  & 5.88  & 5.16  \\
          & 0.4   & 31.13  & 15.52  & 13.12  & 11.18  & 10.59  & 5.87  & 5.16  \\
    \midrule
    \multirow{3}[2]{*}{2} & 0.1   & 121.31  & 28.84  & 17.63  & 14.13  & 12.71  &9.18 & 6.81  \\
          \rowcolor{black!10} 2 & 0.2   & 79.36  & 22.99  & 17.52  & 14.05  & 12.60  & 9.14  & 6.78  \\
          & 0.4   & 54.48  & 22.96  & 17.50  & 14.05  & 12.60  & 9.14  & 6.77  \\
    \bottomrule
    \end{tabular}%
  \label{x_ppl}%
\end{table}%

\subsection{The choice of iterations \textit{it}}
\label{appendix_iter}
\begin{figure}[H]
\begin{center}
\centerline{\includegraphics[width=1\linewidth]{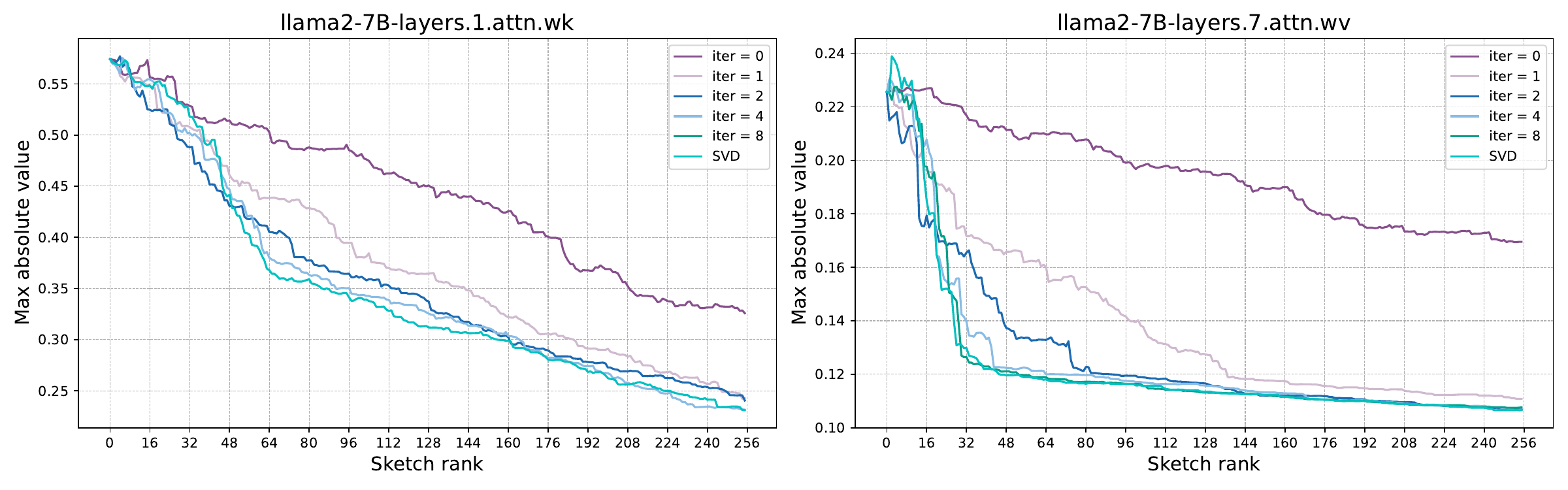}}
\caption{Llama2-7B results}
\label{svd_abs_llama2_7}
\end{center}
\vskip -0.4in
\end{figure}

\begin{figure}[H]
\begin{center}
\centerline{\includegraphics[width=1\linewidth]{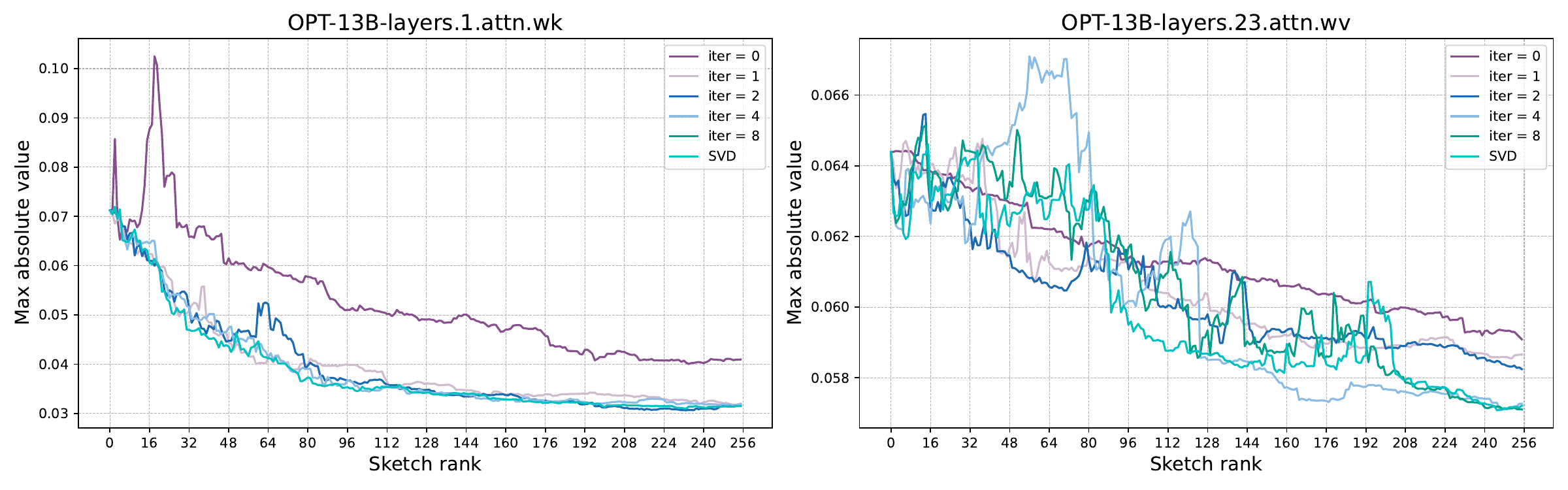}}
\caption{OPT-13B results}
\label{svd_abs_opt_13}
\vskip -0.4in
\end{center}

\end{figure}

In addition to the PPL in the main experimental section, the effect of the $it$ parameter can be observed. A critical basis for rank selection in the FLRQ algorithm is the $absmax$ value of the matrix. To provide a clear visualization of the effects of the R\textbf{1-Sketch} algorithm at various iteration parameters, we also show the variation curve of the absmax extraction effect of the weight matrix in the LLaMA2 and OPT model family with $it$ parameters.
The results are presented from Fig \ref{svd_abs_llama2_7} to Fig \ref{svd_abs_opt_13}, which can also demonstrate the statement in main experiment, $it$ can converge completely when it is around 2, and the approximate accuracy is close to the SVD method.

\textbf{Visualization of different $it$ at $absmax$ in different ranks:}

\begin{figure}[H]
\begin{center}
\centerline{\includegraphics[width=1\linewidth]{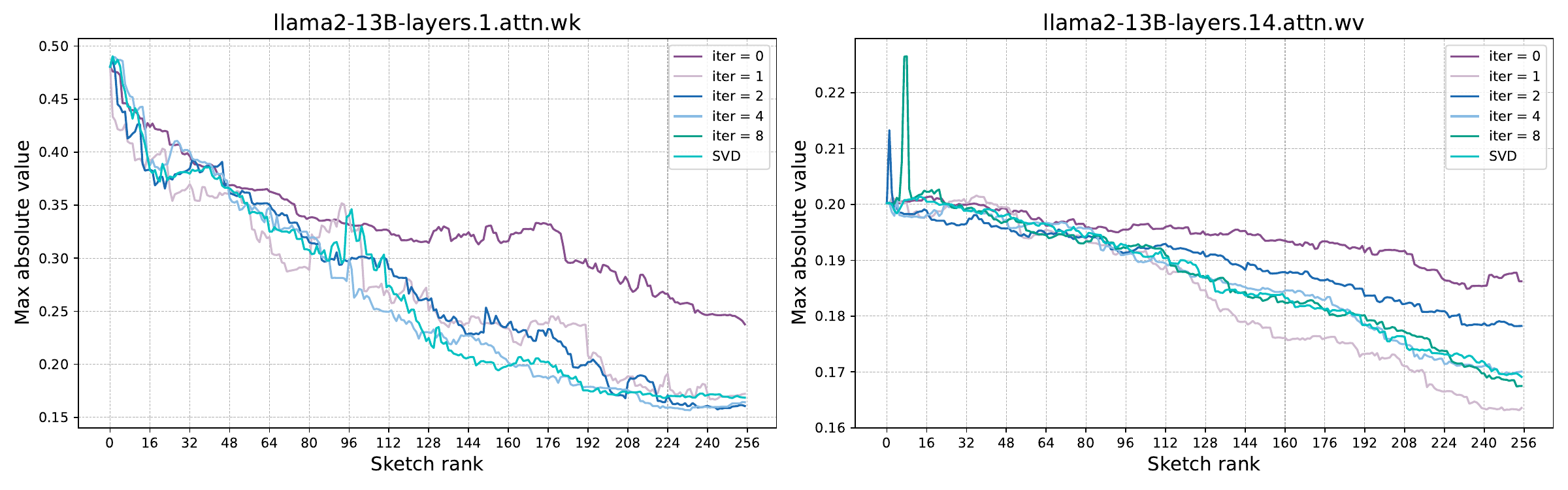}}
\caption{Llama2-13B results}
\label{svd_abs_llama2_13}
\end{center}
\vskip -0.4in
\end{figure}

\begin{figure}[H]

\begin{center}
\centerline{\includegraphics[width=1\linewidth]{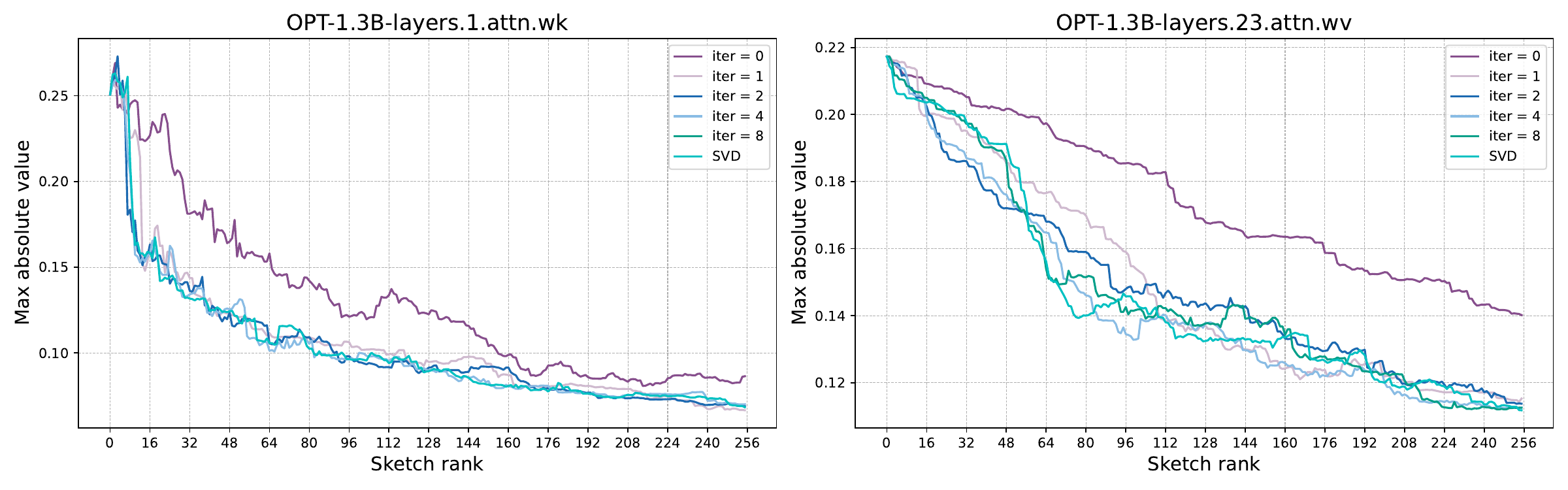}}
\caption{OPT-1.3B results}
\label{svd_abs_opt_1.3}
\end{center}
\vskip -0.2in
\end{figure}

\begin{figure}[H]
\begin{center}
\centerline{\includegraphics[width=1\linewidth]{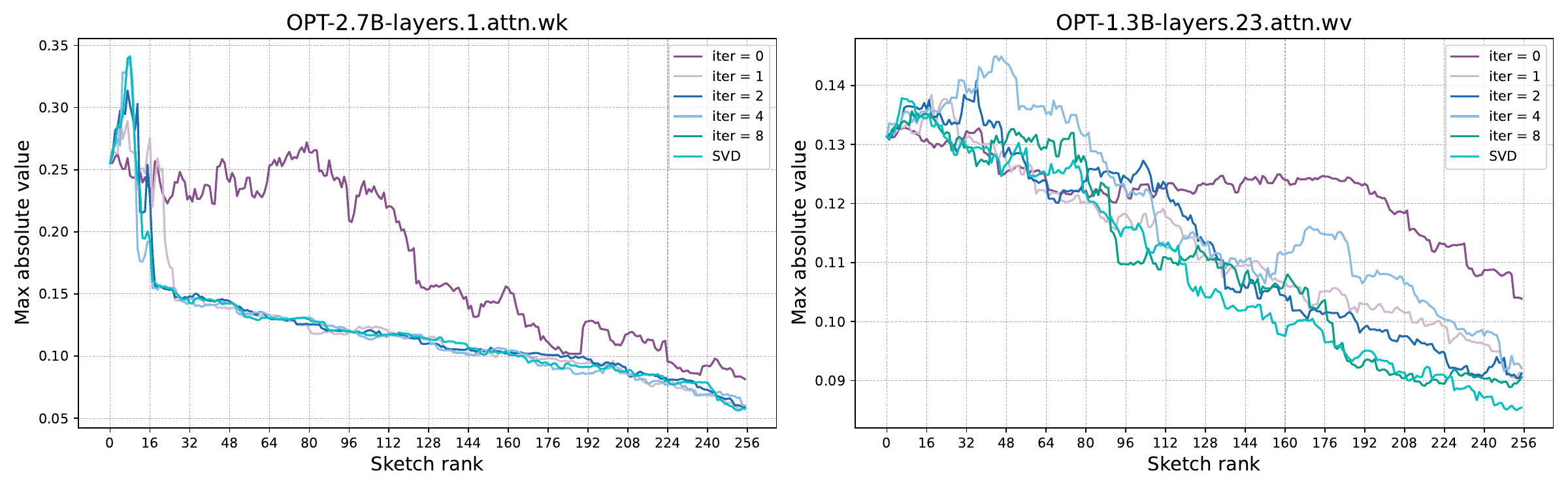}}
\caption{OPT-2.7B results}
\label{svd_abs_opt_2.7}
\end{center}
\vskip -0.2in
\end{figure}

\begin{figure}[H]
\begin{center}
\centerline{\includegraphics[width=1\linewidth]{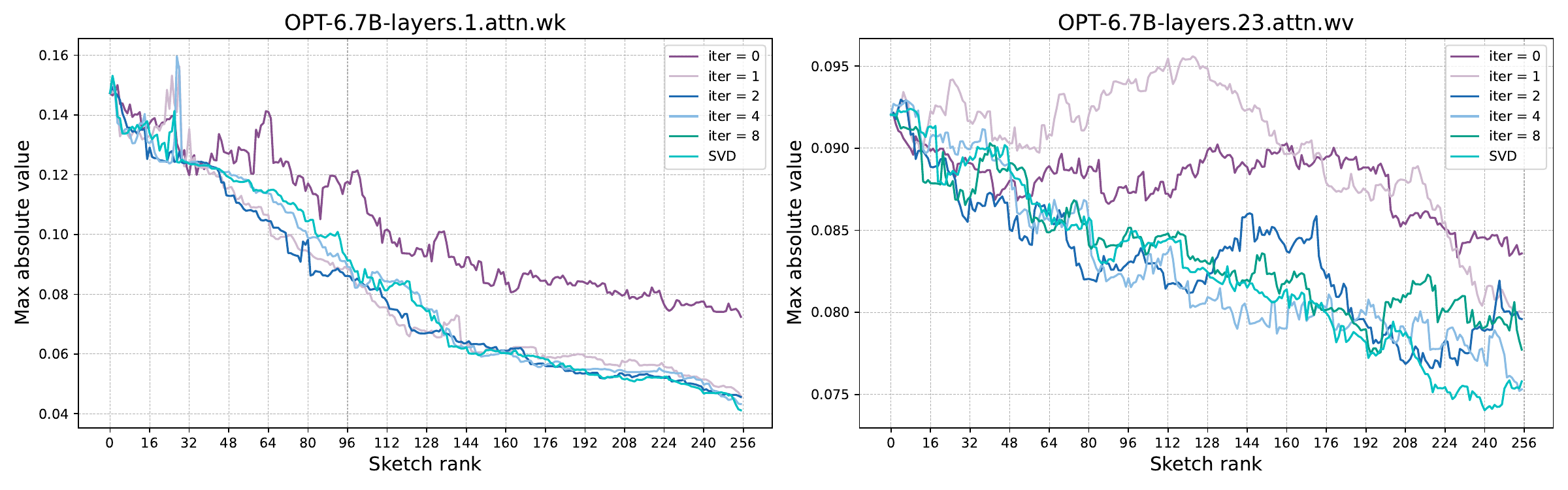}}
\caption{OPT-6.7B results}
\label{svd_abs_opt_6.7}
\vskip -0.4in
\end{center}

\end{figure}

\subsection{The epoch of BLC}
\label{appendix_epoch}

\begin{figure}[H]
\begin{center}
\centerline{\includegraphics[width=1\linewidth]{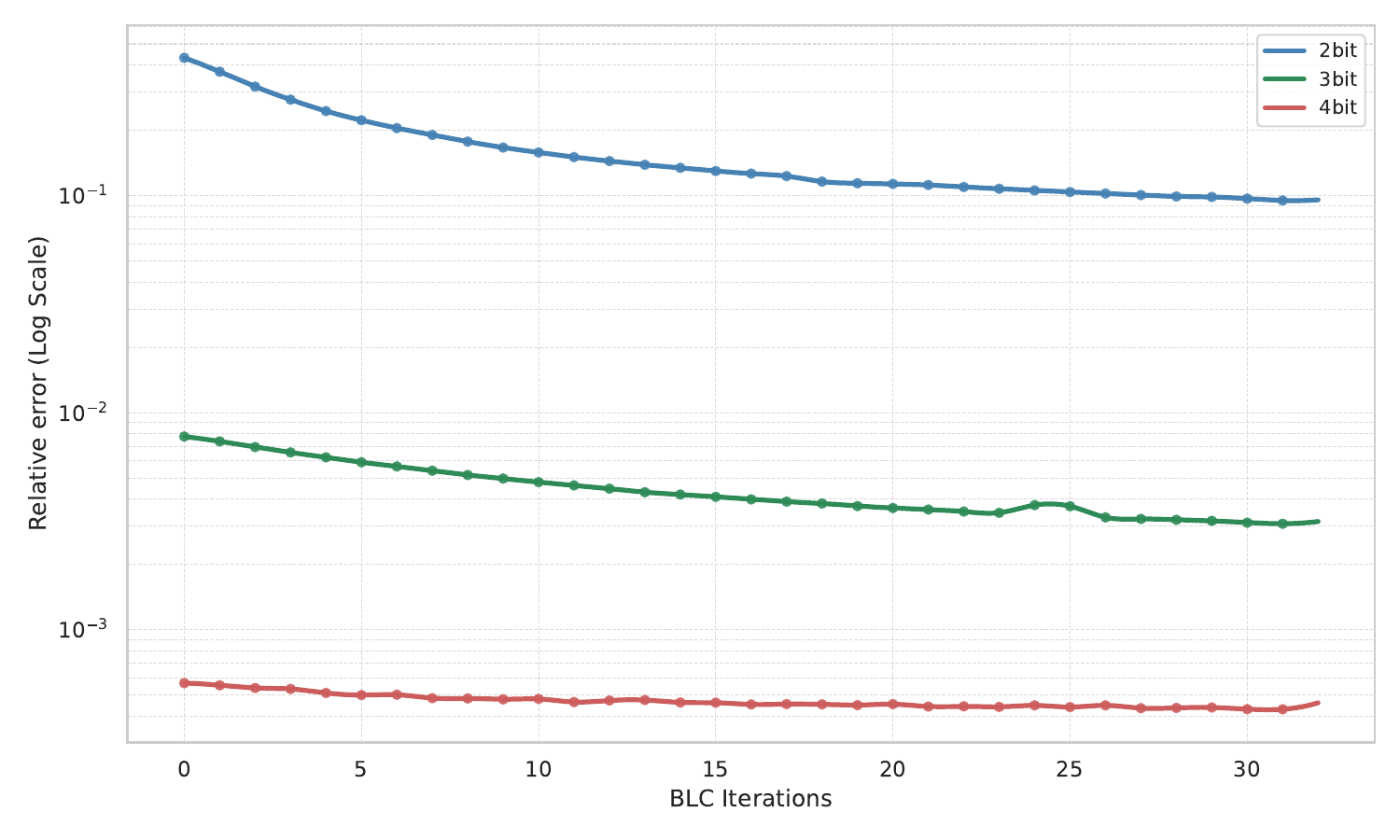}}
\caption{The error curves during BLC iterations at different bit levels.}
\label{blc_iter_error}
\vskip -0.4in
\end{center}
\end{figure}

Since BLC reduces quantization error by alternately updating $W_r$ and $W_q$ in iterative steps, the final accuracy of the algorithm is related to the number of epochs. A larger number of epochs corresponds to higher accuracy but also increases quantization time. The error reduction curve with respect to iterations is shown in Figure \ref{blc_iter_error}, where the error reduction is significantly more pronounced at 2-bit, decreasing by approximately an order of magnitude within 32 epochs, while at 3-bit and 4-bit, the reduction is at most twofold. From the overall model perspective of PPL in Table \ref{tab_blc_ppl}, this corresponds to the BLC algorithm converging within 1 iteration at 3-bit and 4-bit, whereas at 2-bit, approximately 20 iterations are required to achieve optimal accuracy. This also aligns with the results in the quantization time table in the main text, indicating that 2-bit requires more quantization time.

\begin{table}[htbp]
  \centering
  \caption{OPT-6.7b Wiki2 PPL at different BLC.}
    \begin{tabular}{cccccc}
    \toprule
    \multirow{2}[4]{*}{Bit} & \multicolumn{5}{c}{Epoch} \\
\cmidrule{2-6}          & 1     & 5     & 10    & 20    & 30 \\
    \midrule
    4     & \textbf{10.84 } & 10.84  & 10.84  & 10.84  & 10.84  \\
    3     & \textbf{11.18 } & 11.18 & 11.17 & 11.17 & 11.17 \\
    2     & 16.88 & 15.13 & 14.09 & \textbf{14.05 } & 14.05  \\
    \bottomrule
    \end{tabular}%
  \label{tab_blc_ppl}%
\end{table}%

\end{document}